\documentclass[11pt]{article}
\usepackage{fontspec}
\defaultfontfeatures{Ligatures=TeX}
\usepackage{geometry}
\usepackage{graphicx}
\usepackage{booktabs}
\usepackage{float}
\usepackage{xurl}
\usepackage{array}
\usepackage{needspace}
\usepackage{caption}
\usepackage[hidelinks]{hyperref}
\hypersetup{
  pdftitle={When LLMs Sit Above Diagnostic Tools: Unrealized Complementarity in Industrial Fault Diagnosis},
  pdfauthor={Donghwan Kim}
}
\begin{document}
\begin{center}
{\LARGE When LLMs Sit Above Diagnostic Tools: Unrealized Complementarity in Industrial Fault Diagnosis}\\[1.2em]
{\large\bfseries Donghwan Kim}\\[0.3em]
{\large Independent Researcher}\\
{\small\href{mailto:donhkim9714@korea.ac.kr}{donhkim9714@korea.ac.kr}}
\end{center}
\vspace{0.8em}
\section*{Abstract}

Large language models are increasingly used as integration layers above specialized tools, but a stronger component does not necessarily produce a stronger combined system. Across five diagnostic datasets spanning bearing vibration, process monitoring, and semiconductor equipment, we study whether an LLM can reliably use external diagnostic information. Paired repeat calls separate advice effects from ordinary output instability. Conflicting external diagnostic information overturned initially correct LLM judgments in all five datasets; among the four that also provide direct integration comparisons, none showed a consistent advantage for implicit LLM integration over the stronger standalone source. On a Tennessee Eastman confirmation set whose protocol was fixed before evaluation, unaided accuracy was 64.67\%, implicit LLM–specialist integration reached 77.43\%, and the specialist alone reached 83.33\%. Specialist information improved the LLM by 12.8 percentage points (95\% interval 9.7 to 15.9), yet the integrated output remained 5.9 points below the specialist (95\% interval −12.0 to −0.7). A two-source selector oracle reached 92.76\%, indicating case-level complementarity between the standalone predictions that the integrated output did not fully realize. Missed corrections (140 of 295 correction opportunities) far outnumbered harmful acceptances (15 of 99). The deficit remained under prompt and specialist sensitivity analyses. Task-aligned physical evidence reduced susceptibility to incorrect advice in CWRU, whereas higher reasoning effort did not clearly resolve the TEP integration problem. Source quality and integration quality should therefore be evaluated separately: an LLM integration layer should be compared with its stronger standalone component, not only with the unaided LLM.

\section*{1. Introduction}

Large language models need not replace specialized diagnostic models to influence industrial decisions. A vibration classifier or a process monitor can remain the component that turns sensors into evidence. The LLM can instead combine outputs from these components. One context can hold a tool output, a sensor summary, an alarm, a maintenance record, a passage from a manual, and a technician’s note. The model may then prioritize cases, resolve competing hypotheses, or recommend an inspection. Process-systems research already proposes language models as supervisory or coordination layers above numerical tools, retrieved engineering knowledge, and maintenance evidence (\hyperref[cite:gopaluni2026]{Gopaluni, Kotamraju, and Bhushan, 2026}; \hyperref[cite:odonncha2026]{O’Donncha et al., 2026}; \hyperref[cite:liang2026]{Liang and Sin, 2026}; \hyperref[cite:yang2026b]{Yang et al., 2026b}). Once a model performs that role, it directly affects the final decision even though it is not the primary detector. We test whether such a layer improves the final diagnostic decision in controlled settings rather than in deployed plant systems.

Combining informative sources does not necessarily improve the final decision; effective integration requires case-level arbitration between them. Two sources are complementary when each is correct on some cases the other misses. Realizing that complementarity requires choosing appropriately between them. Source quality describes how well each source performs alone. Integration quality describes how well the procedure combines them into a final label. A strong specialist and a competent model can still produce a worse answer than the specialist alone if the integrator ignores a correction or copies an error. Our central question is whether an LLM integrator can turn complementary but imperfect sources into a decision that matches or exceeds the stronger source. The same question applies to other LLM agents that must choose which imperfect source to trust.

Prior work has documented advice following, tool over-use and under-use, and human–AI teams that underperform their stronger member. This study examines those behaviors when the language model itself combines two fallible predictive judgments: an evidence-based LLM assessment and an external specialist prediction.

The protocol separates instability from the effect of the external source. For each case, the model first labels the evidence alone, then receives the same prompt again with no new information to measure ordinary output instability, and finally receives one external diagnosis, which may be correct or incorrect. We ask whether the combined decision outperforms the better standalone source and whether incorrect advice changes an initially correct answer more often than ordinary test–retest variability does.

We study this problem across five diagnostic datasets spanning three industrial settings. In all five, conflicting external diagnostic information overturned some initially correct LLM judgments (Appendix Figure L1). CWRU, Paderborn, XJTU-SY, and Tennessee Eastman provide direct integration comparisons. Across these four datasets, none showed a consistent advantage for implicit LLM integration over the stronger standalone source. Lam 9600 instead provides a controlled-advice real-equipment check. The Tennessee Eastman study provides the primary held-out evaluation because its protocol and confirmation set were fixed before evaluation.

In the held-out TEP evaluation, specialist information improved the unaided LLM by 12.8 points (95\% interval 9.7 to 15.9), yet the integrated output remained 5.9 points below the specialist (95\% interval −12.0 to −0.7). A two-source selector oracle reached 92.76\%, indicating substantial unrealized case-level complementarity. Missed corrections far outnumbered harmful acceptances.

The deficit remained under an explicit arbitration instruction, when the model was also given its stored unaided diagnosis, and with an alternative random-forest specialist. An exploratory Lam 9600 analysis provides a real-equipment example of advice vulnerability.

Our contributions are threefold. First, we distinguish source quality from integration quality and evaluate an LLM integrator directly against the stronger standalone source, using paired repeat calls to separate advice effects from ordinary output instability. Second, across bearing and process-diagnosis settings, we show that implicit LLM integration does not consistently realize the complementarity available between the standalone sources; the held-out TEP evaluation further identifies missed corrections as the dominant integration failure. Third, we test two interventions (task-aligned physical evidence and higher reasoning effort) and show that improvements in one diagnostic property do not necessarily translate into better source integration or uniformly better error profiles.

These results suggest that an LLM integration layer should be evaluated directly against its stronger standalone component, not only against the unaided LLM.

\section*{2. Related work}

Prior work shows that models can over-rely on or under-use tools, and that a combined system can underperform its better member. We study whether an LLM can benefit when each source is correct on some cases the other misses.

Dialogue studies show that assistants adopt a user’s stated preference and sometimes admit mistakes they did not make (\hyperref[cite:perez2022]{Perez et al., 2023}; \hyperref[cite:sharma2023]{Sharma et al., 2023}). \hyperref[cite:xie2024]{Xie et al. (2024)} find that models can be receptive to coherent external evidence that conflicts with stored knowledge, and also that they continue to favor evidence consistent with that knowledge when both are present. In our setting, a specialist class label is provided alongside machine evidence, and success is accuracy against the better source, not agreement with a user.

Recent work shows that source identity and perceived authority can shape how LLMs resolve conflicting information. \hyperref[cite:kumar2026]{Kumar and Chopra (2026)} find that attribution to a verified source induces more compliance than a matched user endorsement and that compliance increases with stated source authority. \hyperref[cite:schuster2026]{Schuster, Gautam, and Markert (2026)} report preferences among source types in inter-context conflicts, while \hyperref[cite:li2025authority]{Li et al. (2025)} find that models can favor user-provided claims over retrieved database evidence when the two conflict. These results indicate that source reliance depends on the setting and source type rather than following a universal authority ordering. These studies characterize source preference or compliance under conflict; we instead ask whether an LLM integrator can benefit when each source is correct on some cases the other misses, and match or exceed the stronger standalone source. For this setting, measuring only adoption of an incorrect source is insufficient because integration can also fail by ignoring a correct specialist when the LLM is wrong.

Tool evaluations already separate adopting a bad return from failing to use a needed one. \hyperref[cite:yang2026a]{Yang et al. (2026a)} corrupt search, delegated answers, and code. Mean adoption of the corrupted return is high for every tool; models sometimes notice the conflict and still return the corrupted answer. \hyperref[cite:soni2026]{Soni (2026)} scores a skipped tool, an ignored result, a fabricated addition, and an unnecessary call as different failures. \hyperref[cite:xu2026]{Xu et al. (2026b)} define “Tool Ignored” as a tool-integrated math failure in which a correct code-execution result is overridden by the model’s own incorrect reasoning. Their calibration method uses confidence in the generated code to decide whether to trust the executor. In our setting, the external source is a fitted diagnostic model with its own error rate, so the integrator must avoid both ignoring correct labels and adopting incorrect ones. We measure under-use where the language model is the integrator and ask whether it can benefit when each source is correct on some cases the other misses.

Recent work has also framed tool use and retrieval as arbitration between fallible sources. MemToC evaluates conflicts between model-specific closed-book answers and executable-tool returns under controlled source correctness (\hyperref[cite:varlamov2026]{Varlamov et al., 2026}), while TrustMargin selects between parametric-memory and retrieval-based answers using a training-free arbitration score (\hyperref[cite:xu2026b]{Xu et al., 2026a}). Our setting differs in both source type and evaluation target: the LLM diagnoses from industrial sensor or process evidence while the second source is a fitted diagnostic specialist whose errors arise naturally. We additionally control for ordinary output instability with paired repeat calls and evaluate whether prompt-level integration exceeds the stronger standalone component.

Human studies ask a person to accept or reject advice. \hyperref[cite:schemmer2023]{Schemmer et al. (2023)} separate relative reliance on correct AI advice from relative self-reliance when the advice is wrong. \hyperref[cite:bansal2021]{Bansal et al. (2021)} ask whether explanations help a human–AI team beat both the human and the model alone, and find that explanations can increase acceptance regardless of correctness. \hyperref[cite:vaccaro2024]{Vaccaro, Almaatouq, and Malone (2024)} find that combinations perform worse than the better partner on average, especially on decisions. In contrast, we study settings in which the language model itself performs the integration.

Learning-to-defer and triage methods explicitly learn or optimize who handles each case (\hyperref[cite:madras2018]{Madras, Pitassi, and Zemel, 2018}; \hyperref[cite:mozannar2020]{Mozannar and Sontag, 2020}; \hyperref[cite:okati2021]{Okati, De, and Gomez-Rodriguez, 2021}; \hyperref[cite:mozannar2023]{Mozannar et al., 2023}). Our setting differs in the mechanism evaluated: we study an off-the-shelf LLM used directly as the integration layer, without a separately trained routing objective.

A similar distinction appears in ensemble studies. \hyperref[cite:kim2026a]{Kim (2026a)} finds oracle gains across language-model subsets and realized majority-vote gains in only a small fraction of them. \hyperref[cite:kim2026b]{Kim (2026b)} finds that a reference-free judge often fails to select a better table-recognition candidate than the first draft. The improvement possible case by case can be much larger than the improvement actually delivered by an aggregation rule. Unlike LLM ensemble settings, our second source is a non-LLM diagnostic model, and the LLM performs case-level integration rather than majority voting.

Industrial fault-diagnosis studies also show that how evidence is presented can matter. \hyperref[cite:lee2025]{Lee et al. (2025)} compare raw sensor windows with descriptive statistics and single- with multi-model diagnostic architectures. \hyperref[cite:li2025s2s]{Li and Zhao (2025)} convert industrial time series into textual trends, periodicity, and deviations before retrieval-guided zero-shot diagnosis. Those studies evaluate diagnostic accuracy and explanation. Our evidence-representation result instead asks whether a model that can already solve a case becomes less likely to abandon that answer after incorrect advice.

Other industrial systems place language models above tools or heterogeneous evidence. FaultExplainer explains principal-component alarms on the Tennessee Eastman process (\hyperref[cite:khan2025]{Khan et al., 2025}). \hyperref[cite:liang2026]{Liang and Sin (2026)} use a language model to coordinate convolutional and ensemble diagnostic tools on the same process and evaluate diagnosis, tool adherence, and report quality. KDACL combines process data, retrieved mechanistic knowledge, and iterative verification to construct causal chains on Tennessee Eastman and a slag-mill system (\hyperref[cite:yang2026b]{Yang et al., 2026b}). AgentRCA pairs a tool-using model with a model of normal dynamics to compare fault hypotheses (\hyperref[cite:wei2026]{Wei and Fink, 2026}), while \hyperref[cite:odonncha2026]{O’Donncha et al. (2026)} combine maintenance records, operational summaries, and failure knowledge in a verification-constrained decision-support layer. These systems establish the practical relevance of the integration-layer role. They mainly evaluate pipeline diagnosis, reporting, causal-chain quality, or controlled deployment. We instead ask whether the final label retains the stronger source’s performance while benefiting from cases where the other source is correct.

Recent industrial LLM benchmarks likewise evaluate capability rather than integration quality. AssetOpsBench benchmarks multi-agent task automation for industrial asset operations and maintenance (\hyperref[cite:patel2026assetops]{Patel et al., 2026}). FactoryBench evaluates machine understanding over industrial robotic telemetry across causal reasoning levels (\hyperref[cite:merzouki2026]{Merzouki et al., 2026}). FailureSensorIQ probes sensor--failure relationships and robustness under perturbation (\hyperref[cite:constantinides2025]{Constantinides et al., 2025}). Those resources ask whether an LLM can solve industrial problems or use industrial tools. Our protocol instead holds two imperfect predictions fixed and measures the integration gap relative to the stronger standalone source; we therefore do not treat those benchmarks as drop-in substitutes for the present evaluation.

\section*{3. Experimental framework}

\subsection*{3.1 Experimental objective}

The experiment evaluates whether an LLM can appropriately arbitrate between two imperfect sources at the case level. It does not ask the model to replace the specialist detector (Figure 1). Source quality is the accuracy of each source alone. Integration quality is the accuracy of the procedure that produces the final label. We define a two-source selector oracle that is correct whenever either stored standalone prediction is correct. It represents the maximum accuracy attainable by selecting between these two stored labels and is not an implemented integration method.

\begin{figure}[H]

\centering

\includegraphics[width=0.98\textwidth]{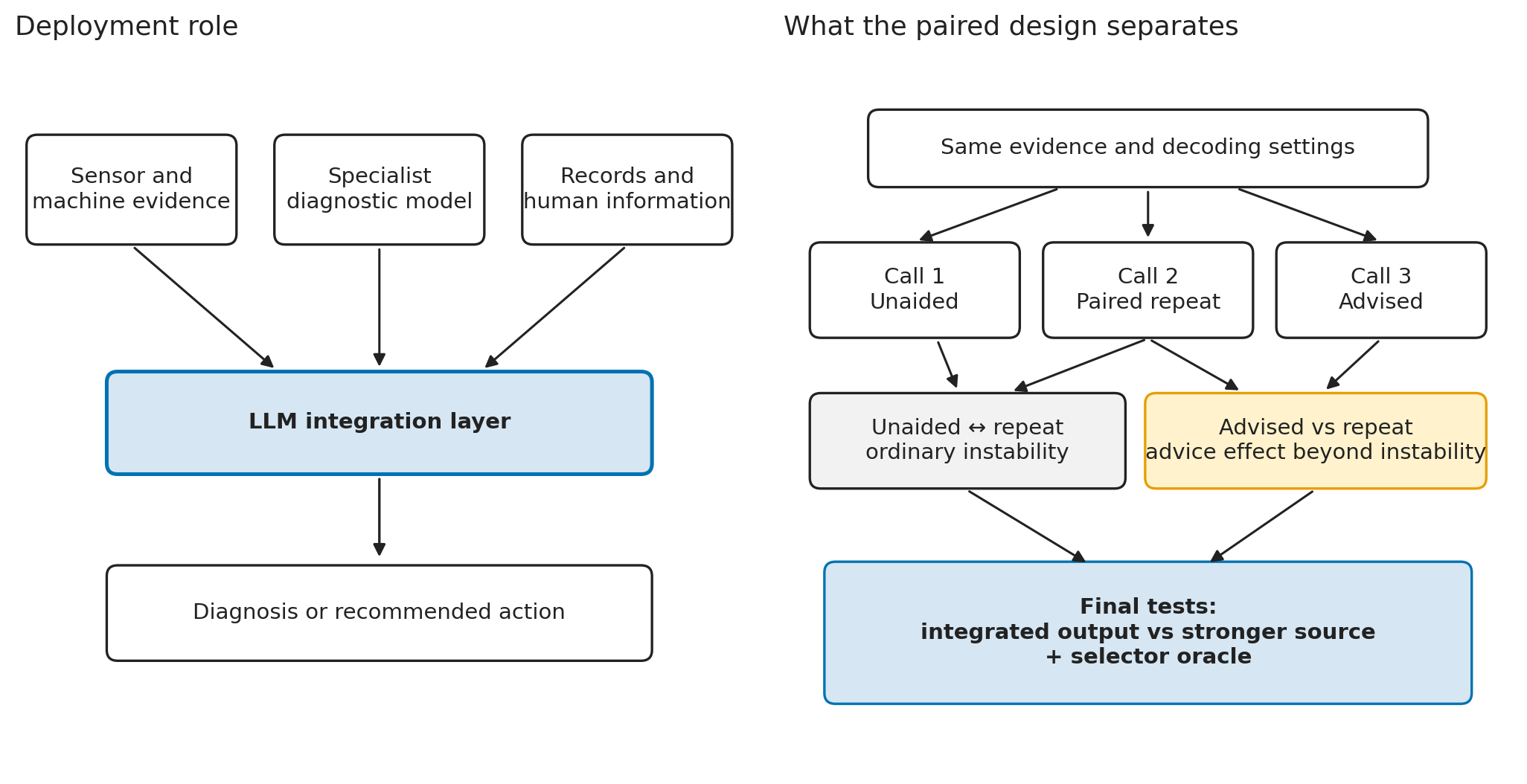}

\caption{The language model is an integration layer, not a replacement detector. Left: machine evidence, a specialist model, and records or human notes can enter one context, and the model may prioritize cases or recommend an action. Right: the paired repeat measures ordinary test–retest instability, while the advised call measures the effect of an external diagnosis beyond that baseline.}

\end{figure}

Each case has a true class and a block of evidence that does not contain the label. The model returns one class from a fixed list. Three calls use that evidence, with the same decoding settings. The first is unaided. The second is an independent repeat with no new information. The repeat is a separate API call with an identical prompt and identical decoding settings; it captures provider-side nondeterminism even at temperature 0. That paired repeat is necessary because a label can change on its own; without it, ordinary instability would be mistaken for the effect of the external source. The third call adds one sentence stating an external diagnostic label. We refer to this advised output as prompt-level implicit integration, or simply implicit integration below: the LLM receives both the case evidence and the external diagnostic prediction in context, without a separately trained routing or fusion mechanism. We then ask whether it outperforms the better of the two standalone sources and whether incorrect advice changes an initially correct judgment beyond ordinary repeat instability. Prompt templates and examples are given in Appendix C.

\subsection*{3.2 Datasets and experimental roles}

The central empirical scope covers five diagnostic datasets across three settings: three bearing datasets, Tennessee Eastman, and Lam 9600. The datasets serve distinct experimental roles rather than matched replications. Table 1 states the role of each. Bearing recordings are where the integration failure was characterized and where a change to the evidence representation was tested. The Tennessee Eastman confirmation protocol was fixed before evaluation and serves as the primary held-out non-bearing integration evaluation; Lam 9600 is an exploratory real-equipment check of advice vulnerability.

\Needspace{18\baselineskip}Table 1. Experimental roles. Bearing datasets characterize source integration and evidence presentation. The Tennessee Eastman set is the held-out non-bearing integration evaluation. Lam 9600 provides an exploratory real-equipment advice-vulnerability check. They are not matched replications.

\begin{table}[H]\centering\scriptsize\setlength{\tabcolsep}{3pt}

\begin{tabular}{p{0.68in}p{0.72in}p{0.90in}p{1.10in}p{0.88in}p{0.67in}p{0.78in}}

\toprule

Dataset & Domain & Task in this study & Evidence shown to the model & External diagnosis & Cluster & Role \\

\midrule

CWRU & Bearing vibration & 4 classes, 72 cases, 28 files & Generic spectral summary, or a task-aligned presentation with additional fault-specific information & Controlled advice, and a fixed classical specialist in the integration comparisons & Recording file & Characterization and intervention \\

Paderborn & Bearing vibration & 3 classes, 87 segments, 29 bearings & Generic and task-aligned text presentations & Controlled advice and fixed fitted specialists & Bearing & Same-domain support \\

XJTU-SY & Bearing run-to-failure & 3 classes, 88 recordings, 11 bearings & Generic and task-aligned text presentations & Controlled advice and fixed fitted specialists & Bearing & Same-domain support \\

Tennessee Eastman & Simulated chemical plant & 6 classes, 150 confirmation runs, 25 per class & Numerical process summary plus a static domain reference (variable roles, subsystems, candidate fault definitions), identical across runs and conditions & Fixed logistic-regression specialist; alternative random-forest specialist in one secondary analysis & Simulation run & Held-out TEP evaluation; secondary sensitivity analyses \\

Lam 9600 & Public metal-etch data from a commercial tool & 5 classes, 20 cases & Tool and trace text & Controlled advice in fixed wording & Wafer & Limited real-equipment check (§4.7; Appendix J) \\

\bottomrule

\end{tabular}

\end{table}

CWRU is the Case Western Reserve University seeded-fault collection (\hyperref[cite:cwru_bearing]{Bearing Data Center}). Paderborn is the electromechanical-drive benchmark of \hyperref[cite:lessmeier2016]{Lessmeier et al. (2016)}. XJTU-SY is the accelerated-life collection released with \hyperref[cite:wang2020]{Wang et al. (2020)}. Process runs come from the \hyperref[cite:rieth2017]{Rieth et al. (2017)} simulation of the \hyperref[cite:downs1993]{Downs and Vogel (1993)} plant. The six TEP classes were fixed before any LLM calls on this dataset and were retained unchanged for confirmation. The Lam 9600 metal-etch data are publicly available from Eigenvector Research and were originally analyzed by \hyperref[cite:wise1999]{Wise et al. (1999)}; we used the machine-state file but do not redistribute the source files. Incorrect labels follow dataset-specific deterministic maps (Appendix Table C1). Bearing integration uses six 400-tree random forests, one strong and one weak predefined feature set for each of CWRU, Paderborn, and XJTU-SY. Strong and weak name those feature sets, not target accuracies. The predictions are out of fold: leave-one-recording-file-out on CWRU and leave-one-bearing-out on Paderborn and XJTU-SY (Appendix G).

\subsection*{3.3 Models}

The fixed panel is seven models, kept unchanged across paired calls for each model–condition combination: GPT-5.6 Sol, Claude Sonnet 5, Claude Opus 5.5, Gemini 3.8 Flash, Qwen 3.8, GLM-5.3, and Kimi K3. Provider routing identifiers and settings are in Appendix A. The CWRU reasoning-effort comparison in Section 4.6 uses five of these models on the same cases and is not pooled with the other two. A separate TEP sensitivity analysis uses four of the seven models; the selected four-model panel was determined before any confirmation outcome at the high reasoning-effort setting was examined.

\subsection*{3.4 Two kinds of external diagnosis}

The external diagnosis is presented in fixed wording, but its source differs across experimental settings.

In the bearing susceptibility and intervention experiments, correctness is set by the design. The external diagnosis is written to be correct, or written to be incorrect, so that a change after incorrect advice can be compared with the paired repeat. Those are controlled advice conditions. Their purpose is to isolate the effect of a conflicting source. Incorrect labels follow the dataset-specific deterministic maps in Appendix Table C1.

Lam likewise uses controlled advice rather than predictions from a standalone diagnostic model. It therefore tests advice vulnerability, not source complementarity or integration-versus-stronger-source performance.

In the integration experiments, including the held-out TEP evaluation and the bearing integration comparisons, the external label is the actual prediction of a fixed specialist model. On the process confirmation set, the specialist is a multinomial logistic regression fit on training runs only. It was selected on development data from two predefined candidates using prespecified eligibility and ranking criteria. The criteria required enough errors and source disagreement to study case-level integration; without such cases, the sources offer little information about when either one can add value. Confirmation outcomes were not used. Appendix Table A1 reports the candidate values and full rule. In the fitted-specialist experiments, errors arise naturally from the specialist model rather than being experimentally assigned. In the fitted-specialist prompts, the specialist prediction was introduced as a case-specific prior and explicitly marked as not itself physical evidence (Appendix C.2). Bearing integration comparisons use the six out-of-fold random forests specified in Section 3.2. This setting tests whether language-model integration benefits when each source is correct on some cases the other misses. Because the source wording and authority differ between controlled-advice and fitted-specialist experiments, we do not compare their absolute switching rates.

\subsection*{3.5 Prompt and specialist sensitivity analyses}

We evaluate three post-primary sensitivity analyses on the same 150 confirmation runs, seven LLMs, unaided outputs, and paired-repeat outputs. The prior-label condition was specified after the arbitration and alternative-specialist results were known.

The arbitration condition retained the selected logistic-regression specialist and changed only the advised prompt by adding one direct arbitration instruction:

{\scriptsize\setlength{\baselineskip}{9.5pt}

\begin{flushleft}\ttfamily

You have two fallible sources: your own reading of the case\par

evidence above and the external diagnostic model's prediction.\par

Decide which diagnosis is better supported by the case evidence.\par

Do not automatically defer to either source.\par

\end{flushleft}}

We treat this prompt as a sensitivity condition rather than an optimized or calibrated fusion procedure. The prior-label condition repeats that arbitration instruction. The specialist prediction remains the preceding case-specific line, \texttt{An external diagnostic advisor suggests: <specialist>.} The inserted line is placed after the first arbitration sentence and immediately before the decision instruction: \texttt{Your earlier diagnosis from the case evidence alone: <z0>.} That order was fixed for all cases and was not randomized. The alternative-specialist condition retained the primary implicit-integration interface but replaced the selected logistic specialist with the alternative 400-tree random forest. That random forest was one of the two development candidates, passed the same eligibility criteria, and was not selected for the primary test under the prespecified ranking rule. Full protocol details and results are in Appendix K. A separate sensitivity analysis with higher reasoning effort is reported in Section 4.6 and Appendix F.2.

\subsection*{3.6 Evidence representations}

On CWRU every request uses an anonymous case identifier and one final evidence block. The generic text is a raw-spectrum summary: band energy, a dominant peak, time-domain shape statistics, and coarse energy shares around precomputed BPFO, BPFI, and BSF bands. It does not provide the bearing geometry, the characteristic frequencies themselves, envelope peaks or harmonics, or an explicit mapping from those frequencies to candidate faults. The task-aligned view adds information absent from the generic representation, including the envelope spectrum at the characteristic frequencies, the theoretical frequency values, and explicit mappings between frequencies and candidate faults. The two views therefore do not carry the same information. Advice is presented in fixed wording, and the model, decoding settings, class list, and response schema are held fixed across paired conditions. The common-correct analysis in Section 4.5 compares cases the model already diagnosed correctly under both views.

Box 1. Illustrative CWRU evidence excerpts (same case, C-000492CE).

\begin{table}[H]\centering\footnotesize\setlength{\tabcolsep}{3pt}

\begin{tabular}{>{\raggedright\arraybackslash}p{3.05in}>{\raggedright\arraybackslash}p{3.05in}}

\toprule

Generic excerpt & Task-aligned excerpt \\

\midrule

- motor load: 0 HP & - BPFO is the outer-race ball-pass frequency. Outer-race defects typically raise envelope energy at BPFO and its harmonics. \\

- rotational speed: 1796 RPM & - BPFI is the inner-race ball-pass frequency. Inner-race defects typically raise envelope energy at BPFI and its harmonics. \\

E1. Spectral band-energy shares: 0-500Hz=0.2\%, 500-1500Hz=0.6\%, 1500-3000Hz=25.3\%, 3000-6000Hz=73.9\%. Spectral entropy=0.647 (+11.1\% vs normal). & - theoretical BPFO=107.3 Hz, BPFI=162.1 Hz, BSF=70.5 Hz, FTF=11.9 Hz \\

E2. Skewness=0.083; impulse factor=8.66 (+111.5\% vs normal). & E1. Envelope spectrum near BPFO (theoretical 107.3 Hz): peak SNR=23.52× local background (nearest peak 108.4 Hz, error 1.1 Hz); 2×BPFO harmonic SNR=19.31×. BPFO is kinematically associated with outer-race defects. \\

\bottomrule

\end{tabular}

\end{table}

The excerpts in Box 1 are from the same case.

Paderborn and XJTU follow the same design principle: a generic spectral summary versus a task-aligned presentation that adds fault-relevant structure. Appendix Table B1 lists the displayed fields for all three bearing datasets.

On the Tennessee Eastman confirmation set, every prompt includes a static domain reference that is identical across runs and conditions and contains no run-specific information: variable roles, subsystem membership and pathway groups, and candidate fault definitions. For the primary TEP integration condition, unaided, repeat, and advised calls differ only by the external diagnosis sentence. The case evidence is a numerical process summary of the fifty-two variables. For the TEP confirmation prompts, that summary is compared with a normal reference constructed from the 50 NORMAL runs in the training split; development and confirmation runs were not used in that construction.

The process-development contrast uses a different intervention. Its aligned view groups the same measurements by subsystem and pathway without adding a new physical transform. We verified that every summary could be computed from the same input data. This development contrast is analyzed separately from the held-out TEP evaluation.

\subsection*{3.7 Inclusion criteria for bearing comparisons}

A bearing comparison enters the count in Section 4.4 only if it meets the following predefined inclusion rules. Unaided accuracy must exceed chance by at least 0.15: 0.40 on the four-class CWRU task and about 0.48 on the three-class bearing tasks. We also exclude comparisons when the accuracy gap between the selector oracle and the better source is under 0.03 or the specialist is incorrect on fewer than 15 cases; there is then too little evidence to estimate the integration-versus-source difference reliably. Twenty-two comparisons meet these rules. The CWRU specialists do not enter that count because each made fewer than 15 errors. All 28 CWRU model–condition estimates (7 models $\times$ 2 evidence presentations $\times$ 2 specialists) were still below the stronger source. The rules came from the bearing verification protocol and were not chosen from the comparisons summarized here. The 22-comparison count is conditional on them.

\subsection*{3.8 Evaluation metrics}

Table 2 defines the quantities used in the results.

\Needspace{18\baselineskip}Table 2. Evaluation metrics.

\begin{table}[H]\centering\footnotesize\setlength{\tabcolsep}{3pt}

\begin{tabular}{p{1.45in}p{2.45in}p{2.15in}}

\toprule

Quantity & Definition & Why it is reported \\

\midrule

Unaided accuracy & Share of cases correct before any external diagnosis & The model’s own reading of the evidence \\

Test–retest instability & How often the paired repeat changes the unaided label & Separates ordinary instability from the effect of advice \\

Retest-adjusted harmful switching & Among initially correct rows, the advised-call error rate minus the repeat-call error rate & The susceptibility the intervention experiments target \\

Integrated output versus the better source & Accuracy of the advised judgment minus the better of unaided accuracy and specialist accuracy & Whether integration outperforms the stronger component \\

Two-source selector oracle & Correct whenever either of the two stored standalone labels is correct & Descriptive bound for selection between the two stored labels \\

Missed correction & The specialist is correct, the unaided LLM is wrong, and the final integrated output remains wrong & Failure to use a correct specialist correction \\

Harmful acceptance & The specialist is wrong, the unaided LLM is correct, and the final integrated output becomes wrong & Failure to preserve a correct unaided judgment \\

\bottomrule

\end{tabular}

\end{table}

Let \(A^{\mathrm{int}}_m\), \(A^{\mathrm{unaided}}_m\), and \(A^{\mathrm{spec}}_m\) be the integrated, unaided, and specialist accuracies of model \(m\). The headline contrast is the unweighted mean across the \(M\) evaluated models,

\(\Delta_{\mathrm{best}} = \frac{1}{M}\sum_{m=1}^{M}\left[A^{\mathrm{int}}_m - \max\left(A^{\mathrm{unaided}}_m, A^{\mathrm{spec}}_m\right)\right].\)

The better standalone source is selected at the model level, not case by case; the same calculation is recomputed within each joint cluster-bootstrap resample.

Let \(z_0\) be the unaided label, \(z_r\) the repeat label, \(z_a\) the advised label, and \(y\) the true label. The adjusted harmful-switching metric is:

\(H_{\mathrm{adj}} = P(z_a \ne y \mid z_0 = y) - P(z_r \ne y \mid z_0 = y).\)

The denominator is therefore the set of initially correct rows. Positive values mean that the advised call became wrong more often than the repeat call; negative values are possible. Each cluster-bootstrap sample recomputes both terms and their difference.

\subsection*{3.9 Bootstrap confidence intervals}

We compute 95\% percentile bootstrap intervals from 10,000 resamples without adjustment for multiple comparisons. The resampling unit matches each dataset: recording files for CWRU, physical bearings for Paderborn and XJTU, and simulation runs for the process confirmation set. The bootstrap seed is \path{20260924}. The supplementary integration-versus-unaided interval jointly resamples the 150 process runs across the fixed seven-model panel. The post hoc CWRU panel summary jointly resamples all 28 recording files across the same fixed panel; models are not resampled.

\section*{4. Results}

\subsection*{4.1 Implicit LLM integration remains below the specialist}

Seven models saw the same generic numerical summary of 150 Tennessee Eastman runs. Unaided accuracy was 64.67\%, implicit integration reached 77.43\%, and the specialist alone reached 83.33\% (Figure 2). The gain of the integrated output over the unaided models was 12.8 points, with a supplementary run-bootstrap interval from 9.7 to 15.9. The primary contrast, integrated output minus specialist, is −5.9 points, with a 95\% interval from −12.0 to −0.7. Because the specialist was at least as accurate as every unaided model, this contrast is also $\Delta_{\mathrm{best}}$ as defined in Section 3.8. Differences use the unrounded seven-model means. One model matched the specialist both unaided and after advice, so the average does not imply that every model changed in the same direction.

\begin{figure}[H]

\centering

\includegraphics[width=0.98\textwidth]{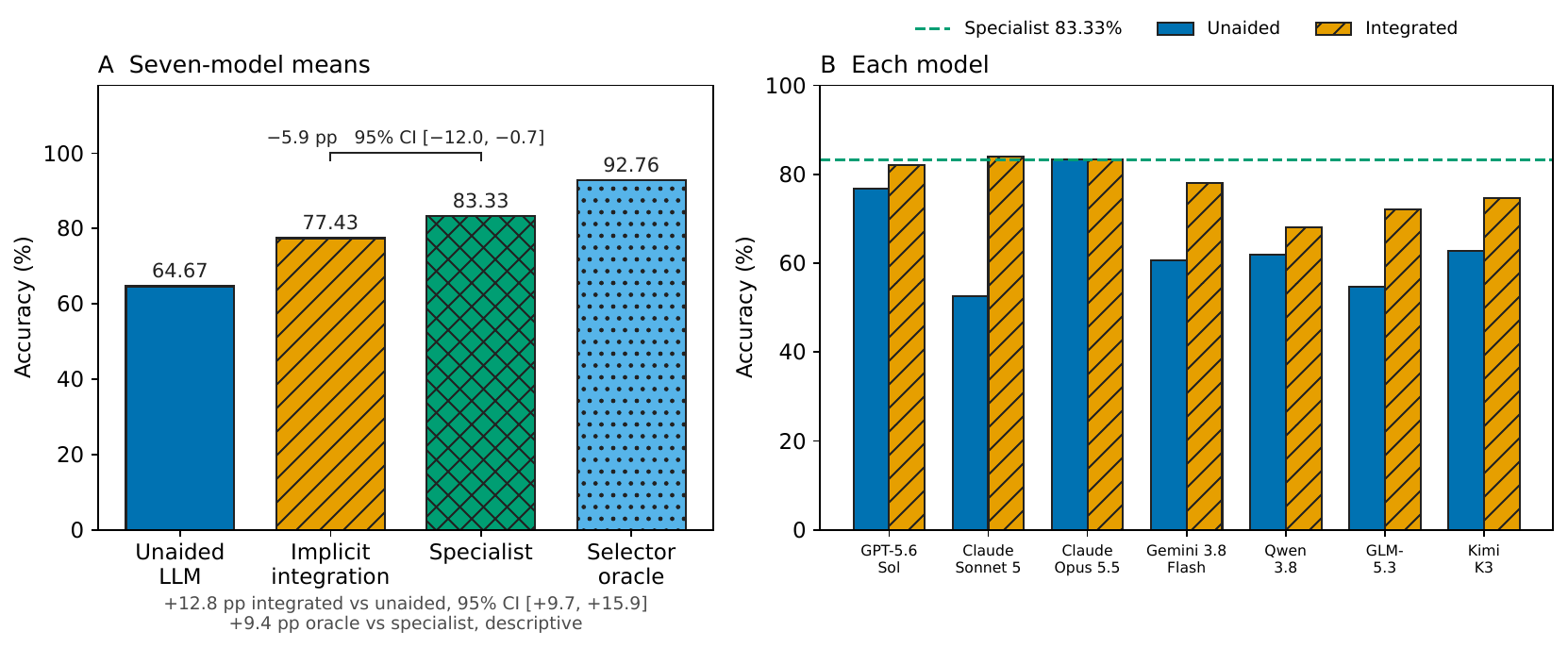}

\caption{Specialist information substantially improved the language model, yet implicit LLM integration remained below the specialist alone. Panel A: unaided 64.67\%, integrated output 77.43\%, specialist 83.33\%, two-source selector oracle 92.76\%. The integrated output is 5.9 points below the specialist (95\% interval −12.0 to −0.7) and 12.8 points above the unaided model (supplementary 95\% interval 9.7 to 15.9). Differences use unrounded seven-model means. The selector oracle is a descriptive bound on the two stored labels. Panel B: unaided and integrated accuracy for each model, with the specialist selected on development data shown as the reference line. One model matches that line both unaided and after integration.}

\end{figure}

The two-source selector oracle reached 92.76\%, 9.4 points above the specialist, reflecting cases on which each standalone source was correct and the other was not.

\subsection*{4.2 Most integration errors are missed corrections}

The dominant observed failure was not broad adoption of incorrect advice. It was failure to use correct specialist information when the model was wrong.

Figure 3 separates the rows on which the two standalone sources were correct. The sources disagreed in 394 of 1,050 model–case rows: the specialist alone was correct in 295, and the unaided LLM alone was correct in 99. The integrated output missed 140 of the 295 specialist corrections (47.5\%) and lost 15 of the 99 initially correct LLM judgments (15.2\%). Fourteen of those 15 harmful acceptances copied the specialist’s wrong label, all in the normal class. Nine additional errors occurred when both standalone sources were correct, and 73 occurred when both were wrong.

\begin{figure}[H]

\centering

\includegraphics[width=0.98\textwidth]{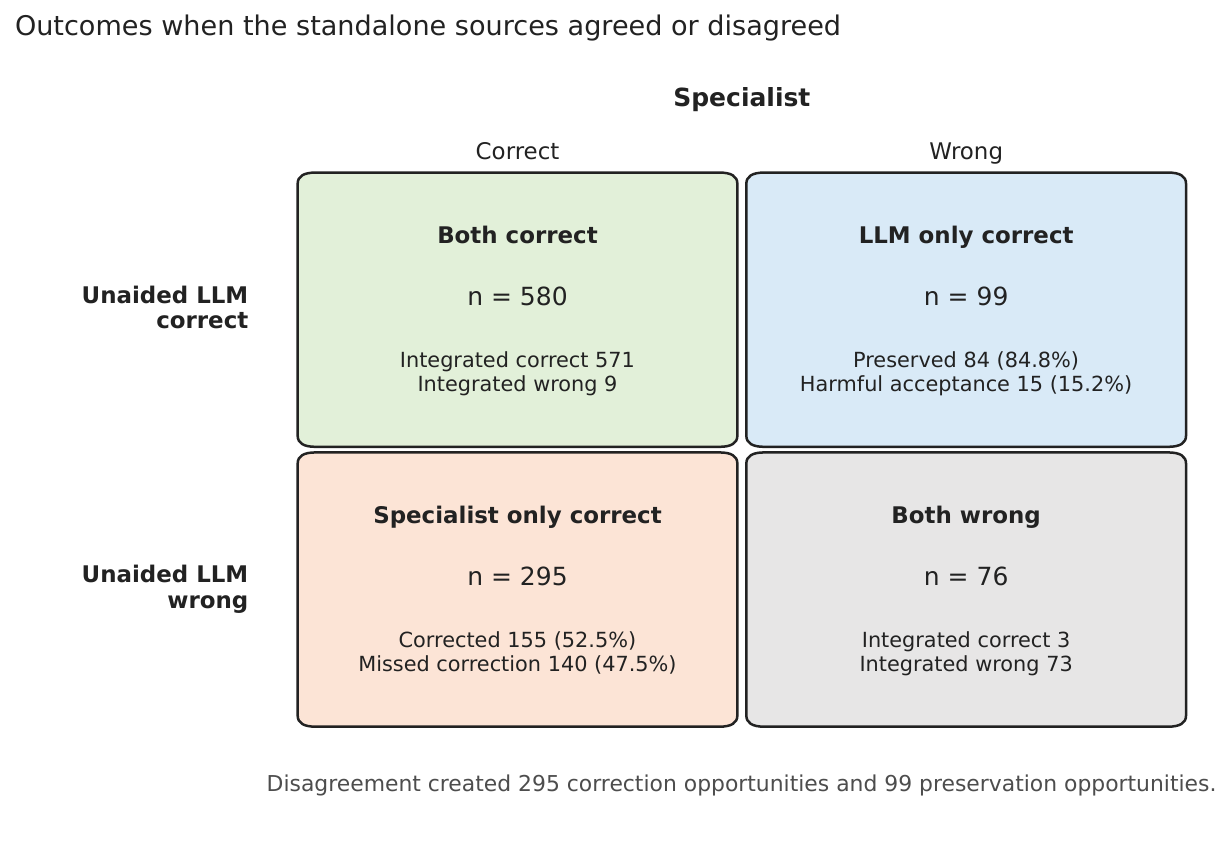}

\caption{Source correctness and final integrated outcomes in the held-out Tennessee Eastman evaluation. The two sources disagreed in 394 rows, creating 295 opportunities to use a specialist correction and 99 opportunities to preserve a correct unaided LLM judgment. The integrated output missed 140 of the 295 corrections (47.5\%) and lost 15 of the 99 correct LLM judgments (15.2\%). Cell counts sum to 1,050 model–case rows. These counts describe error types; they should not be summed to recover the 5.9-point accuracy gap.}

\end{figure}

Box 2. Illustrative TEP decision sequences.

\begin{table}[H]\centering\footnotesize\setlength{\tabcolsep}{3pt}

\begin{tabular}{>{\raggedright\arraybackslash}p{1.25in}>{\raggedright\arraybackslash}p{1.05in}>{\raggedright\arraybackslash}p{0.65in}>{\raggedright\arraybackslash}p{0.65in}>{\raggedright\arraybackslash}p{0.75in}>{\raggedright\arraybackslash}p{0.65in}>{\raggedright\arraybackslash}p{0.75in}}

\toprule

Error type & Model & True class & Unaided & Paired repeat & Specialist & Advised \\

\midrule

\mbox{Missed correction} & Qwen 3.8 & IDV1 & IDV6 & IDV6 & IDV1 & IDV6 \\

\mbox{Harmful acceptance} & Claude Sonnet 5 & NORMAL & NORMAL & NORMAL & IDV15 & IDV15 \\

\bottomrule

\end{tabular}

\end{table}

Box 2 contains illustrative examples only. Examples were selected deterministically as the first stored instance of each error pattern, using fixed case and model order, for which the paired repeat reproduced the unaided label.

The pooled counts hide substantial model-level variation (Figure 4). Relative self-reliance (RSR) is the share of initially correct LLM judgments preserved when the specialist is wrong. Relative AI reliance (RAIR) is the share of correct specialist judgments accepted when the LLM is initially wrong (\hyperref[cite:schemmer2023]{Schemmer et al., 2023}). Claude Sonnet 5 had high RAIR (0.966) but low RSR (0.250), while Claude Opus 5.5 had RSR 1.000 but RAIR 0.000. The other models fell between these extremes. The models showed different source-reliance profiles. No model reached the region that is high on both axes.

\begin{figure}[H]

\centering

\includegraphics[width=0.66\textwidth]{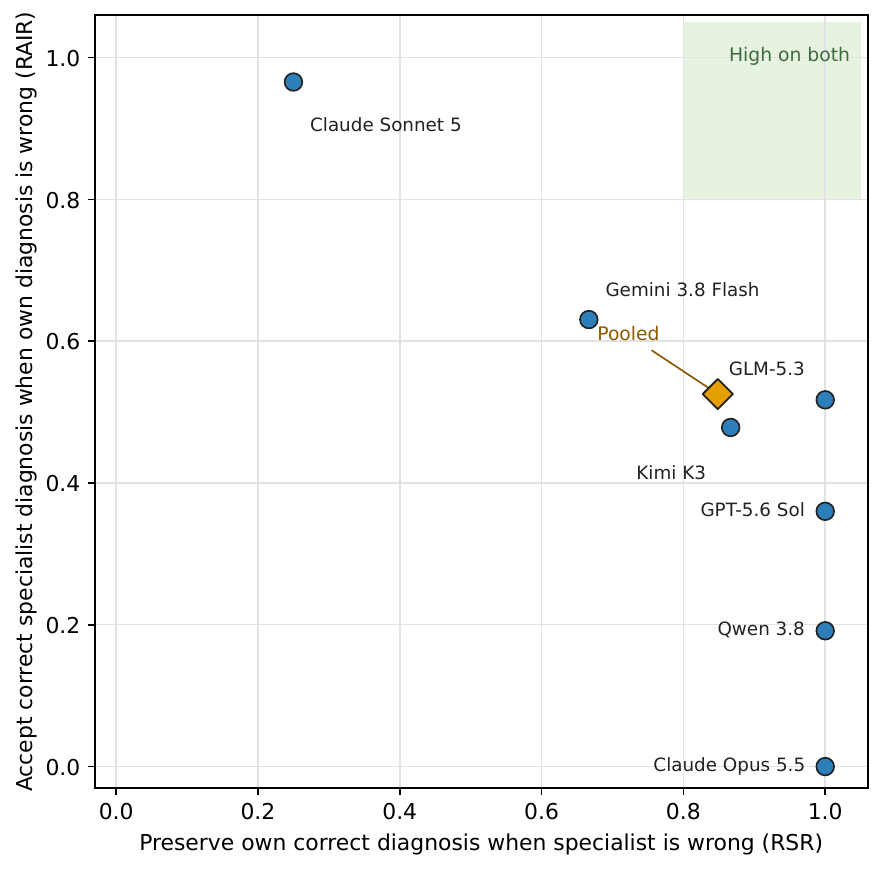}

\caption{Model-level source reliance on the Tennessee Eastman confirmation set. Blue circles are individual models, and the orange diamond pools all model–case opportunities. The pale area marks rates of at least 0.8 on both axes as a visual guide, not a decision threshold. Position shows how a model treats advice, not its overall accuracy. These reliance rates should be interpreted together with each model’s standalone accuracy; Figure 2B shows the resulting integrated accuracies. Descriptive 95\% Wilson intervals and opportunity counts are in Appendix Table A2. The figure does not estimate a population of models.}

\end{figure}

Incorrect advice did affect the final judgment beyond ordinary test–retest instability, but this effect was secondary to the missed corrections. On rows where the unaided label was right and the specialist was wrong, the paired repeat never produced the specialist’s wrong label. Across models, the advised call adopted that exact wrong label 16.4 percentage points more often than the paired repeat (95\% interval 9.5 to 22.6; unweighted model mean). The corresponding pooled exact-copy count was 14 of 99. The effect was concentrated in two models. The specialist’s own mistakes were also concentrated: 14 in the normal class, 10 in one fault class, and 1 in another, out of 25 specialist-wrong runs. A post-hoc analysis of the true-NORMAL runs further showed that the integrator often resisted specialist false alarms when the unaided LLM was correct (Appendix M).

\subsection*{4.3 Prompt and specialist sensitivity analyses of the TEP result}

The primary integration deficit was not removed by a direct source-arbitration instruction, by additionally providing the stored unaided diagnosis, or by the alternative random-forest specialist (Table 3). These are secondary sensitivity analyses, not independent confirmations.

\Needspace{18\baselineskip}Table 3. Sensitivity analyses of the Tennessee Eastman integration deficit.

\begin{table}[H]\centering\scriptsize\setlength{\tabcolsep}{3pt}

\begin{tabular}{>{\raggedright\arraybackslash}p{1.15in}>{\raggedright\arraybackslash}p{0.62in}>{\raggedright\arraybackslash}p{0.58in}>{\raggedright\arraybackslash}p{1.55in}>{\raggedright\arraybackslash}p{0.86in}>{\raggedright\arraybackslash}p{0.86in}}

\toprule

Condition & Specialist accuracy & Integrated accuracy & Integrated − better standalone source (95\% CI) & Missed corrections & Harmful acceptances \\

\midrule

Primary: logistic, implicit & 83.3\% & 77.4\% & −5.9 [−12.0, −0.7] & 140/295 (47.5\%) & 15/99 (15.2\%) \\

Arbitration condition: logistic, explicit instruction & 83.3\% & 76.1\% & −7.2 [−14.0, −1.3] & 161/295 (54.6\%) & 6/99 (6.1\%) \\

Prior-label condition:\newline logistic, stored unaided diagnosis & 83.3\% & 72.2\% & −11.1 [−18.3, −3.6] & 215/295 (72.9\%) & 0/99 (0.0\%) \\

Alternative-specialist condition: random forest, implicit & 84.0\% & 78.2\% & −5.7 [−11.2, −1.1] & 116/272 (42.6\%) & 10/72 (13.9\%) \\

\bottomrule

\end{tabular}

\end{table}

Explicit arbitration did not improve integrated accuracy relative to the implicit condition (explicit minus implicit −1.3 percentage points, 95\% interval −3.1 to +0.5). It reduced harmful acceptances but increased missed corrections, changing the composition of errors without improving overall source selection. Reintroducing the stored unaided diagnosis strengthened this trade-off further (prior-label minus arbitration −3.9 points, 95\% interval −5.8 to −2.0), eliminating harmful acceptances while increasing missed corrections (Appendix K). Replacing the logistic-regression specialist with the alternative random-forest specialist produced the same qualitative integration deficit (−5.7 points versus the better standalone source, 95\% interval −11.2 to −1.1).

\subsection*{4.4 Bearing studies rarely show reliable gains over the stronger source}

Across the included bearing analyses, implicit integration rarely outperformed the stronger standalone source. Of 22 eligible comparisons, three intervals were below zero, eighteen included zero, and one was above zero (Section 3.7; Appendix G). All 28 CWRU point estimates were still below the stronger source. The intervals are not adjusted for multiple comparisons, and the pattern does not imply that every integrated output fell below the stronger source.

\subsection*{4.5 Task-aligned evidence reduces susceptibility to incorrect advice}

We next evaluate task-aligned physical evidence as an intervention for reducing susceptibility to conflicting advice.

Task-aligned evidence reduced susceptibility to incorrect advice across most models in the CWRU common-correct analysis. We measure the extent to which initially correct diagnoses are switched after incorrect advice, adjusting for ordinary paired test–retest instability. The comparison is restricted to cases correctly diagnosed under both evidence presentations.

Six of seven point estimates were positive, and five confidence intervals excluded zero (Figure 5). Estimates ranged from 0.00 to 0.48; model-specific values, intervals, and common-correct sample sizes are shown in Figure 5 and Appendix D. As a post hoc summary across the fixed seven-model panel, the unweighted mean reduction was 0.246 [0.105, 0.405] under a joint file-cluster bootstrap.

\begin{figure}[H]

\centering

\includegraphics[width=0.98\textwidth]{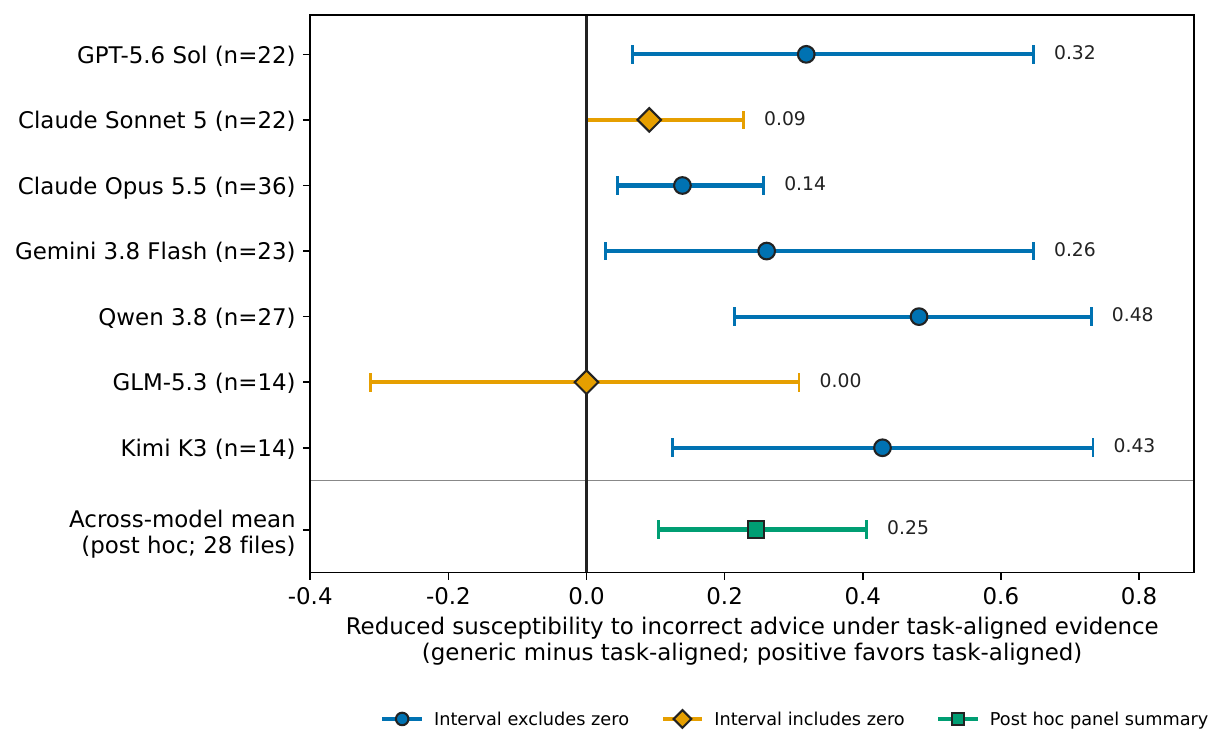}

\caption{In the CWRU common-correct analysis, task-aligned evidence yielded lower harmful-switching estimates for six of the seven evaluated models; five model-specific confidence intervals excluded zero. Points are generic minus task-aligned retest-adjusted harmful switching with 95\% file-cluster intervals. Positive values mean reduced susceptibility under task-aligned evidence. Common-correct cases are those the same model labeled correctly, before advice, under both presentations; model-specific sample sizes are shown. The square is a post hoc unweighted summary across the fixed seven-model panel, with all 28 recording files resampled jointly across models.}

\end{figure}

Paderborn and XJTU provide supporting bearing-domain evidence, with changes in diagnostic competence accompanying the changes in advice-related switching behavior (Appendix H). The same benefit did not appear when the TEP development measurements were grouped into deterministic subsystem summaries (difference −1.7 percentage points, 95\% CI −3.6 to +0.3; Appendix I). The CWRU result therefore supports task-aligned evidence presentation as an intervention that helps in this setting, not a general benefit of adding or reorganizing structure. A post-hoc decomposition further showed that the task-aligned presentations changed the balance between normal/healthy-to-fault and fault-to-normal/healthy errors in a dataset-dependent manner (Appendix N). Susceptibility was also not consistently associated with a separately computed evidence-strength score in CWRU or in the prespecified held-out TEP analysis (Appendix E).

\subsection*{4.6 Higher reasoning effort does not clearly resolve the integration deficit}

We next tested whether the high reasoning-effort setting reduced susceptibility to incorrect advice. On CWRU, it did not yield a statistically reliable reduction for any of the five evaluated models. Neither unaided accuracy nor susceptibility to incorrect advice showed a reliable improvement across those five models: every model-specific interval comparing the low and high settings included zero (Appendix F).

Because the primary held-out TEP evaluation used the low reasoning-effort setting, we additionally evaluate a post-primary higher-reasoning sensitivity analysis. The four-model panel was determined from development diagnostics before any higher-reasoning confirmation outcome was examined (Appendix F.2). Higher reasoning increased unaided accuracy by 5.2 points (95\% interval 2.2 to 8.2), but integrated accuracy increased by only 2.2 points (−0.2 to 4.8). In this four-model subset, integrated minus the better standalone source was −6.3 points at the low setting (−13.5 to 1.0) and −4.2 points at the high setting (−10.5 to 0.8). Higher reasoning improved unaided LLM accuracy but did not provide clear evidence of changing or resolving the integration deficit.

\subsection*{4.7 Exploratory evaluation on real-equipment data}

An exploratory analysis using public Lam 9600 data from a commercial semiconductor etch tool provides a real-equipment example of advice vulnerability. Of 23 initially correct model–case judgments, six were overturned by incorrect external advice across four wafers, while the paired repeats did not flip. This analysis is descriptive and is not intended as a benchmark of diagnostic capability (Appendix J). Controlled labels make this a vulnerability check, not a test of source complementarity.

\section*{5. Discussion}

\subsection*{5.1 Source quality is not integration quality}

The integration problem was not confined to one benchmark. Across the four datasets with direct integration comparisons, implicit integration did not show a consistent advantage over the stronger standalone source. On Paderborn and XJTU-SY, only one of 22 eligible comparison intervals was above zero; all 28 CWRU point estimates were below the stronger source, although specialist errors were too sparse for comparable inference. The held-out TEP evaluation provides the strongest evidence.

A specialist source can make a language model more accurate and still leave the combined system below that specialist. In TEP, the specialist label was clearly useful, yet implicit integration failed to preserve the specialist’s standalone performance despite substantial case-level complementarity indicated by the selector oracle. Because the development selection criteria required enough disagreement to study integration, the oracle gap describes this selected setting rather than what should be expected from an arbitrary specialist.

Before placing an LLM integration layer above a validated specialist, the layer should itself be validated against that source.

The sensitivity analyses changed the balance between missed corrections and harmful acceptance without removing the integration deficit. They do not establish failure for every prompting or fusion procedure, or for every specialist family, but they show that the primary finding is not specific to the original prompt or to the selected logistic-regression specialist.

\subsection*{5.2 Missed corrections dominate harmful acceptance}

Strategies aimed only at limiting adoption of incorrect tool output address harmful acceptance but do not address the missed corrections that dominated the TEP error analysis. Under-use of correct specialist advice was the dominant failure mode. Incorrect advice still had an effect, but it was concentrated in a small number of models and classes. Evaluations that measure only adoption of incorrect specialist outputs would therefore overlook the dominant failure mode.

The explicit arbitration instruction illustrates the same tension: harmful acceptance decreased, but missed corrections increased. A rule that only suppresses tool following therefore need not improve case-level source selection.

The NORMAL-class analysis shows the other side of this trade-off: when the specialist was wrong, resisting it was often the correct action (Appendix M). Thus, the problem is not uniformly too much or too little specialist reliance, but whether reliance is allocated to the correct source on each case.

The missed-correction pattern is consistent with the “Tool Ignored” failure reported for tool-integrated math reasoning (\hyperref[cite:xu2026]{Xu et al., 2026b}), here observed when the external source is itself imperfect. The bearing comparisons did not show a single dominant direction of source-selection failure. In the process experiment, the integrated output often failed to use a correct specialist label, and those rows were common enough to dominate the error count. Across settings, case-level source selection was unreliable. The experiments do not show why models over-use advice in some settings and under-use it in others.

\subsection*{5.3 Interventions affecting susceptibility to incorrect advice}

In our experiments, presenting task-aligned physical evidence reduced susceptibility to incorrect advice for most CWRU models. Reorganizing the same Tennessee Eastman measurements did not produce the same benefit (Section 4.5). Higher reasoning effort did not yield a reliable reduction in that CWRU susceptibility. The TEP sensitivity analysis shows that an improvement in standalone accuracy need not translate into a clear improvement in source integration. Higher reasoning effort therefore should not be assumed to solve case-level source arbitration automatically.

These intervention results motivate a broader question: whether improvement on a targeted metric corresponds to uniformly better diagnostic behavior. The post-hoc error-profile analysis suggests that it need not: evidence representation can shift the balance among error types even when another targeted metric improves (Appendix N).

These results suggest three practical checks: compare the integrated output with the stronger standalone source; examine consequential error types separately rather than relying on aggregate accuracy; and test whether the evidence representation exposes the physical structure needed for diagnosis. Verification-constrained industrial systems show the value of constructing and checking evidence before producing recommendations (\hyperref[cite:odonncha2026]{O’Donncha et al., 2026}; \hyperref[cite:yang2026b]{Yang et al., 2026b}). Our results add that the final system should also be compared directly with its stronger standalone component.

The exploratory Lam 9600 analysis extends the broader advice vulnerability to public data from a commercial semiconductor etch tool, but its controlled-advice design does not provide an integration-versus-source comparison.

\section*{6. Limitations}

The study uses controlled interfaces for external diagnostic information. In the bearing advice experiments, the external diagnosis was presented in fixed wording and its correctness was set by design, so that the effect of a conflicting source could be separated from test–retest instability. In the integration experiments, including the held-out TEP evaluation, the external label came from a fixed specialist model with naturally occurring errors, but it was still presented as a short fixed diagnostic statement rather than a live engineer note or a full tool log. In the fitted-specialist prompts, the specialist label was framed as a case-specific prior and marked as not itself physical evidence. This framing may have encouraged discounting of the specialist relative to the machine evidence; alternative source presentations could shift the balance between missed corrections and harmful acceptance.

The primary TEP condition evaluates implicit integration. Broader structured or learned fusion procedures were not evaluated. The prior-label condition used a fixed prompt order in which the stored unaided diagnosis immediately preceded the decision instruction. The design therefore cannot distinguish anchoring on the prior diagnosis from position or recency effects. The higher-reasoning sensitivity analysis used four of the seven primary models because development diagnostics did not reliably distinguish the requested high setting from low for the others. Manipulation strength remained provider-specific, and the high setting also used a larger output-token limit. The analysis therefore tests provider-side reasoning effort under a concise-output interface rather than explicitly elicited written deliberation.

The primary test used the multinomial logistic specialist selected on development data, while a secondary analysis with the alternative random-forest specialist also showed the integration deficit. Both are classical diagnostic models, so other specialist families may produce different error profiles and integration behavior.

We evaluated source integration across four diagnostic datasets spanning three bearing benchmarks and the Tennessee Eastman process, with an additional real-equipment check of advice vulnerability on Lam 9600. The Lam check uses controlled advice rather than predictions from a standalone fitted specialist and therefore does not test complementarity. Tennessee Eastman provides the primary held-out non-bearing evaluation, but it is a simulated process rather than historical process data from an operating plant. The TEP evaluation uses a predefined six-class subset of the broader fault taxonomy rather than the full set of TEP fault modes. The subset was fixed before any LLM calls on the dataset, but conclusions need not extend to the remaining fault modes. Specialist mistakes and the copying of those mistakes were concentrated in a few classes and models, so the effect varied across classes and models.

The Lam analysis uses few wafers and three models and is exploratory rather than a benchmark evaluation. Some rendered XJTU prompts were not retained, which limits exact prompt-level reproduction of that auxiliary analysis.

The design does not show whether baseline diagnostic accuracy causes models to over- or under-rely on advice. The bearing and process results both show poor case-level source selection, but they do not explain why the direction differs.

\section*{7. Broader impacts}

Industrial diagnostic decisions can affect equipment availability, maintenance, and safety. The LLM layers evaluated here should not be deployed autonomously; our results argue for validating any such layer against the strongest available source.

\section*{8. Conclusion}

Source quality and integration quality are distinct. Across four diagnostic datasets with direct integration comparisons, prompt-level implicit LLM integration did not consistently outperform the stronger standalone source. In the held-out TEP evaluation, specialist information substantially improved the unaided LLM, yet the integrated output remained below the specialist, with missed corrections dominating harmful acceptance. Task-aligned physical evidence reduced susceptibility to incorrect advice in the CWRU setting, whereas higher reasoning effort did not clearly resolve the TEP integration deficit. These results motivate evaluating LLM integration layers directly against their stronger standalone components rather than only against unaided LLMs.

\section*{References}

\phantomsection\label{cite:bansal2021}
Bansal, G., Wu, T., Zhou, J., Fok, R., Nushi, B., Kamar, E., Ribeiro, M. T., and Weld, D. S. (2021). Does the whole exceed its parts? The effect of AI explanations on complementary team performance. \emph{Proceedings of the 2021 CHI Conference on Human Factors in Computing Systems}, Article 81.

\phantomsection\label{cite:cwru_bearing}
Case Western Reserve University. Bearing Data Center. \url{https://engineering.case.edu/bearingdatacenter}

\phantomsection\label{cite:constantinides2025}
Constantinides, C., Patel, D., Lin, S., Guerrero, C., Patil, S. D., and Kalagnanam, J. (2025). FailureSensorIQ: A multi-choice QA dataset for understanding sensor relationships and failure modes. \emph{Advances in Neural Information Processing Systems (NeurIPS) Datasets and Benchmarks Track}. arXiv:2506.03278.

\phantomsection\label{cite:downs1993}
Downs, J. J. and Vogel, E. F. (1993). A plant-wide industrial process control problem. \emph{Computers \& Chemical Engineering}, 17(3), 245–255.

\phantomsection\label{cite:gopaluni2026}
Gopaluni, R. B., Kotamraju, V., and Bhushan, S. (2026). Large language models in process systems engineering: Opportunities, architectures, and industrial deployment challenges. arXiv:2606.11589.

\phantomsection\label{cite:khan2025}
Khan, A., Nahar, R., Chen, H., Constante Flores, G. E., and Li, C. (2025). FaultExplainer: Leveraging large language models for interpretable fault detection and diagnosis. \emph{Computers \& Chemical Engineering}, 199, 109152. \url{https://doi.org/10.1016/j.compchemeng.2025.109152}

\phantomsection\label{cite:kim2026a}
Kim, D. (2026a). Are diversity metrics measuring diversity? A capability-controlled audit of majority-vote gain in LLM ensembles. arXiv:2607.20768.

\phantomsection\label{cite:kim2026b}
Kim, D. (2026b). LLM-as-a-judge scores are unreliable optimization signals in closed-loop table recognition. arXiv:2607.13347.

\phantomsection\label{cite:kumar2026}
Kumar, A. R. and Chopra, P. (2026). Authority Bias in Language Models: Source Deference and User Agreement Are Not Interchangeable. arXiv:2609.37616.

\phantomsection\label{cite:lee2025}
Lee, X. Y., Vidyaratne, L., Farahat, A., and Gupta, C. (2025). Exploring LLM-based agentic frameworks for fault diagnosis. \emph{Annual Conference of the Prognostics and Health Management Society}. \url{https://doi.org/10.36001/phmconf.2025.v17i1.4350}.

\phantomsection\label{cite:lessmeier2016}
Lessmeier, C., Kimotho, J. K., Zimmer, D., and Sextro, W. (2016). Condition monitoring of bearing damage in electromechanical drive systems by using motor current signals of electric motors: A benchmark data set for data-driven classification. \emph{Proceedings of the European Conference of the Prognostics and Health Management Society}, 3(1). \url{https://doi.org/10.36001/phme.2016.v3i1.1577}

\phantomsection\label{cite:li2025s2s}
Li, B. and Zhao, C. (2025). S2S-FDD: Bridging industrial time series and natural language for explainable zero-shot fault diagnosis. \emph{2025 CAA Symposium on Fault Detection, Supervision, and Safety for Technical Processes}, 1–6. \url{https://doi.org/10.1109/SAFEPROCESS67117.2025.11268252}

\phantomsection\label{cite:li2025authority}
Li, Y., Guo, X., Gao, J., Chen, G., Zhao, X., Zhang, J., Liu, Q., Wu, H., Yao, X., and Wei, X. (2025). LLMs Trust Humans More, That’s a Problem! Unveiling and Mitigating the Authority Bias in Retrieval-Augmented Generation. \emph{Proceedings of the 63rd Annual Meeting of the Association for Computational Linguistics (Volume 1: Long Papers)}, 28844–28858. \url{https://doi.org/10.18653/v1/2025.acl-long.1400}

\phantomsection\label{cite:liang2026}
Liang, J. and Sin, G. (2026). A large language model enhanced fault diagnosis framework for chemical processes. \emph{Systems and Control Transactions}, 5, 1769–1775. \url{https://doi.org/10.69997/sct.118894}

\phantomsection\label{cite:madras2018}
Madras, D., Pitassi, T., and Zemel, R. (2018). Predict responsibly: Improving fairness and accuracy by learning to defer. \emph{Advances in Neural Information Processing Systems}, 31. \url{https://papers.nips.cc/paper_files/paper/2018/hash/09d37c08f7b129e96277388757530c72-Abstract.html}

\phantomsection\label{cite:merzouki2026}
Merzouki, Y., Izquierdo, C., Ignuta-Ciuncanu, M., G{\'o}mez-Bracamonte, M., Maggioni, R., Lombardi, A., Mazzoleni, C., Martelli, F., G{\"u}nther, B., Petersen, J., and Petersen, P. (2026). FactoryBench: Evaluating industrial machine understanding. arXiv:2605.07675.

\phantomsection\label{cite:mozannar2020}
Mozannar, H. and Sontag, D. (2020). Consistent estimators for learning to defer to an expert. \emph{Proceedings of the 37th International Conference on Machine Learning}, PMLR 119, 7076–7087. \url{https://proceedings.mlr.press/v119/mozannar20b.html}

\phantomsection\label{cite:mozannar2023}
Mozannar, H., Lang, H., Wei, D., Sattigeri, P., Das, S., and Sontag, D. (2023). Who Should Predict? Exact Algorithms For Learning to Defer to Humans. \emph{Proceedings of the 26th International Conference on Artificial Intelligence and Statistics}, PMLR 206, 10520–10545. \url{https://proceedings.mlr.press/v206/mozannar23a.html}

\phantomsection\label{cite:odonncha2026}
O’Donncha, F., Zhou, N., Martinez, N., Rayfield, J. T., Heath, F. F. III, Langbridge, A., and Vaculin, R. (2026). Evidence-driven reasoning for industrial maintenance using heterogeneous data. \emph{Proceedings of the 64th Annual Meeting of the Association for Computational Linguistics (Volume 6: Industry Track)}, 736–748. \url{https://doi.org/10.18653/v1/2026.acl-industry.50}

\phantomsection\label{cite:okati2021}
Okati, N., De, A., and Gomez-Rodriguez, M. (2021). Differentiable learning under triage. \emph{Advances in Neural Information Processing Systems}, 34. \url{https://proceedings.neurips.cc/paper_files/paper/2021/hash/4c4c937b67cc8d785cea1e42ccea185c-Abstract.html}

\phantomsection\label{cite:patel2026assetops}
Patel, D., Lin, S., Rayfield, J., Zhou, N., Shyalika, C., Yarrabothula, S. R., Vaculin, R., Martinez, N., O’Donncha, F., and Kalagnanam, J. (2026). AssetOpsBench: A real-world evaluation benchmark for AI-driven task automation in industrial asset management. \emph{Proceedings of the 32nd ACM SIGKDD Conference on Knowledge Discovery and Data Mining (KDD ’26)}, 9649–9660. \url{https://doi.org/10.1145/3770855.3817562}. arXiv:2506.03828.

\phantomsection\label{cite:perez2022}
Perez, E., Ringer, S., Lukošiūtė, K., Nguyen, K., Chen, E., Heiner, S., Pettit, C., Olsson, C., Kundu, S., Kadavath, S., et al. (2023). Discovering language model behaviors with model-written evaluations. \emph{Findings of the Association for Computational Linguistics: ACL 2023}, 13387–13434. \url{https://doi.org/10.18653/v1/2023.findings-acl.847}.

\phantomsection\label{cite:rieth2017}
Rieth, C. A., Amsel, B. D., Tran, R., and Cook, M. B. (2017). Additional Tennessee Eastman process simulation data for anomaly detection evaluation. Harvard Dataverse. \url{https://doi.org/10.7910/DVN/6C3JR1}

\phantomsection\label{cite:schemmer2023}
Schemmer, M., Kühl, N., Benz, C., Bartos, A., and Satzger, G. (2023). Appropriate reliance on AI advice: Conceptualization and the effect of explanations. \emph{Proceedings of the 28th International Conference on Intelligent User Interfaces}. arXiv:2302.02187.

\phantomsection\label{cite:schuster2026}
Schuster, J., Gautam, V., and Markert, K. (2026). Whose Facts Win? LLM Source Preferences under Knowledge Conflicts. \emph{Proceedings of the 64th Annual Meeting of the Association for Computational Linguistics (Volume 1: Long Papers)}, 29430–29459. \url{https://doi.org/10.18653/v1/2026.acl-long.1357}

\phantomsection\label{cite:sharma2023}
Sharma, M., Tong, M., Korbak, T., Duvenaud, D., Askell, A., Bowman, S. R., Cheng, N., Durmus, E., Hatfield-Dodds, Z., Johnston, S. R., et al. (2023). Towards understanding sycophancy in language models. arXiv:2310.13548.

\phantomsection\label{cite:soni2026}
Soni, H. (2026). ToolFailBench: Diagnosing tool-use failures in LLM agents. arXiv:2607.04686.

\phantomsection\label{cite:vaccaro2024}
Vaccaro, M., Almaatouq, A., and Malone, T. W. (2024). When combinations of humans and AI are useful: A systematic review and meta-analysis. \emph{Nature Human Behaviour}, 8, 2293–2303.

\phantomsection\label{cite:varlamov2026}
Varlamov, A., Zinnatullin, R., Rykov, E., Panchenko, A., and Alimova, I. (2026). MemToC: Benchmarking memory-tool conflict resolution in large language models. arXiv:2608.26295.

\phantomsection\label{cite:wang2020}
Wang, B., Lei, Y., Li, N., and Li, N. (2020). A hybrid prognostics approach for estimating remaining useful life of rolling element bearings. \emph{IEEE Transactions on Reliability}, 69(1), 401–412.

\phantomsection\label{cite:wei2026}
Wei, A. and Fink, O. (2026). Agentic root cause analysis through evidence-grounded reasoning. arXiv:2607.22385.

\phantomsection\label{cite:wise1999}
Wise, B. M., Gallagher, N. B., Butler, S. W., White, D. D., Jr., and Barna, G. G. (1999). A comparison of principal component analysis, multiway principal component analysis, trilinear decomposition and parallel factor analysis for fault detection in a semiconductor etch process. \emph{Journal of Chemometrics}, 13(3–4), 379–396. \url{https://doi.org/10.1002/(SICI)1099-128X(199905/08)13:3/4<379::AID-CEM556>3.0.CO;2-N}

\phantomsection\label{cite:xie2024}
Xie, J., Zhang, K., Chen, J., Lou, R., and Su, Y. (2024). Adaptive chameleon or stubborn sloth: Revealing the behavior of large language models in knowledge conflicts. \emph{The Twelfth International Conference on Learning Representations (ICLR 2024).} \url{https://openreview.net/forum?id=auKAUJZMO6}

\phantomsection\label{cite:xu2026b}
Xu, J., Shi, H., Shan, Y., Liu, P., Bai, Y., Li, N., and Liu, X. (2026a). TrustMargin: Training-free arbitration between parametric memory and retrieved evidence in large language models. arXiv:2606.08397.

\phantomsection\label{cite:xu2026}
Xu, R., Ji, Y., Luo, Y., Li, J., Li, D., Li, P., Li, J., and Zhang, M. (2026b). When to trust tools? Adaptive tool trust calibration for tool-integrated math reasoning. \emph{Findings of the Association for Computational Linguistics: ACL 2026}, 17691–17706. \url{https://doi.org/10.18653/v1/2026.findings-acl.877}

\phantomsection\label{cite:yang2026a}
Yang, H., Song, W., Kim, T., Song, J., Park, S., and Jo, Y. (2026a). Agents’ overreliance on unreliable tools. arXiv:2609.05587.

\phantomsection\label{cite:yang2026b}
Yang, T., Zhu, M., Ji, Y., Li, Q., and Chen, B. (2026b). KDACL: A knowledge-data fusion-driven method for interpretable anomaly causal chain construction in process industries using large language models. \emph{Journal of Intelligent Manufacturing}. \url{https://doi.org/10.1007/s10845-026-02830-6}

\section*{Appendix A. Reproducibility details}

The TEP confirmation protocol and specialist-selection procedure were fixed before confirmation evaluation. Analyses use fixed stored outputs and the dataset-appropriate cluster bootstrap procedures defined in Section 3.9 (seed \path{20260924}). All LLM calls were issued through the OpenRouter API with a single model identifier and a pinned upstream provider (provider.order); provider fallback was disabled, and no model-fallback models array was used. Decoding settings for each prompt family are reported in Appendix C.6. Across the paper, the OpenRouter model identifiers were \path{openai/gpt-5.6-sol}, \path{anthropic/claude-sonnet-5}, \path{anthropic/claude-opus-5.5}, \path{google/gemini-3.8-flash}, \path{qwen/qwen3.8-2.4t-a95b}, \path{z-ai/glm-5.3}, and \path{moonshotai/kimi-k3}.

Two specialist candidates were fixed before selection: a 400-tree random forest and multinomial logistic regression (\path{C=1.0}) with standardized inputs. Each was fit on 300 training runs, 50 per class, and evaluated on the same 150-run development set.

\Needspace{18\baselineskip}\emph{Appendix Table A1. Development-only TEP specialist selection.}

\begin{table}[H]\centering\scriptsize\setlength{\tabcolsep}{3pt}

\begin{tabular}{p{1.35in}p{0.65in}p{0.5in}p{1.2in}p{1.35in}p{0.65in}}

\toprule

Candidate & Dev. accuracy & Errors & Mean selector-oracle gap & Pooled LLM-only-correct rows & Decision \\

\midrule

Random forest & 0.820 & 27 & 0.074 & 80 & Passed \\

Multinomial logistic regression & 0.833 & 25 & 0.096 & 101 & Selected \\

\bottomrule

\end{tabular}

\end{table}

The prespecified eligibility criteria required at least 15 specialist errors, a mean selector-oracle gap of at least 0.03 across models, at least 15 pooled LLM-only-correct rows, and at least one model with unaided accuracy at or above specialist accuracy. Both candidates passed and each had one such model. The fixed ranking then preferred the larger pooled LLM-only-correct count, followed by the larger gap if needed. It selected logistic regression. The criteria had no separate minimum accuracy threshold. Confirmation outcomes were not used for selection.

Figure 3 is a descriptive cross-tabulation of all 1,050 Tennessee Eastman model–case rows by unaided-LLM correctness and specialist correctness. No row was excluded from that matrix, and no additional inference was performed.

\Needspace{18\baselineskip}\emph{Appendix Table A2. Descriptive model-level reliance on the TEP confirmation set.}

\begin{table}[H]\centering\scriptsize\setlength{\tabcolsep}{3pt}

\begin{tabular}{p{0.85in}p{0.95in}p{0.95in}p{1.35in}p{1.45in}}

\toprule

Model & RSR [95\% Wilson interval] & RSR n & RAIR [95\% Wilson interval] & RAIR n \\

\midrule

GPT-5.6 Sol & 1.000 [0.796, 1.000] & 15 & 0.360 [0.202, 0.555] & 25 \\

Claude Sonnet 5 & 0.250 [0.089, 0.532] & 12 & 0.966 [0.883, 0.990] & 58 \\

Claude Opus 5.5 & 1.000 [0.796, 1.000] & 15 & 0.000 [0.000, 0.204] & 15 \\

Gemini 3.8 Flash & 0.667 [0.391, 0.862] & 12 & 0.630 [0.486, 0.755] & 46 \\

Qwen 3.8 & 1.000 [0.796, 1.000] & 15 & 0.191 [0.104, 0.325] & 47 \\

GLM-5.3 & 1.000 [0.796, 1.000] & 15 & 0.517 [0.392, 0.641] & 58 \\

Kimi K3 & 0.867 [0.621, 0.963] & 15 & 0.478 [0.341, 0.619] & 46 \\

\bottomrule

\end{tabular}

\end{table}

The model-level opportunities in Appendix Table A2 sum to the Figure 3 totals: 99 preservation opportunities and 295 correction opportunities.

\section*{Appendix B. Dataset details}

The CWRU task contains 72 four-class cases clustered in 28 recording files. Paderborn contains 87 segments clustered in 29 physical bearings. XJTU-SY contains 88 recordings clustered in 11 bearings. The Tennessee Eastman confirmation set contains 150 simulation runs, 25 each from NORMAL, IDV1, IDV4, IDV6, IDV13, and IDV15. The Lam analysis uses the publicly available Eigenvector Lam 9600 machine-state data described by \hyperref[cite:wise1999]{Wise et al. (1999)}.

\Needspace{18\baselineskip}\emph{Appendix Table B1. Fields displayed in the generic and task-aligned bearing presentations.}

\begin{table}[H]\centering\scriptsize\setlength{\tabcolsep}{3pt}

\begin{tabular}{p{0.75in}p{2.55in}p{2.85in}}

\toprule

Bearing dataset & Generic presentation & Task-aligned additions or substitutions \\

\midrule

CWRU & Load, speed, sensor, and window metadata; broad spectral-band shares; spectral entropy; skewness; impulse factor; RMS and normal-reference change; coarse energy shares around precomputed BPFO, BPFI, and BSF bands, without the frequencies or fault mapping; kurtosis; dominant component & Generic diagnostic reference; geometry-derived BPFO, BPFI, BSF, and FTF frequencies; envelope peak and harmonic signal-to-background ratios and offsets; matched-normal RMS and impulsiveness comparisons \\

Paderborn & Speed; RMS, kurtosis, crest factor, peak-to-peak, and skewness; ordinary-spectrum dominant component and relative power; spectral entropy; band-energy fractions & Documented bearing geometry and characteristic-frequency equations; BPFO, BPFI, BSF, and FTF values; envelope-analysis band; local peak frequency, offset, and signal-to-background ratio; deterministic BPFO–BPFI contrasts \\

XJTU-SY & Speed; RMS, kurtosis, crest factor, peak-to-peak, and skewness; ordinary-spectrum dominant component and relative power; spectral entropy; band-energy fractions & Documented bearing geometry; BPFO, BPFI, BSF, and FTF values; envelope-analysis band; local peak frequency, offset, and signal-to-background ratio; deterministic BPFO–BPFI contrast \\

\bottomrule

\end{tabular}

\end{table}

Controlled-advice analyses set the external diagnosis to be correct or incorrect and use fixed presentation wording. Integration analyses instead use predictions from a specialist model fitted before evaluation, whose errors occur naturally. These settings answer different questions and are not pooled.

\section*{Appendix C. Prompts}

This appendix reports the prompt families used in the paper. The paired-repeat prompt matches the unaided prompt byte-for-byte and is omitted.

\subsection*{C.1 Shared instructions and output format}

System-level output-format instruction, where applicable:

{\scriptsize\setlength{\baselineskip}{9.5pt}

\begin{flushleft}\ttfamily

Reply with valid JSON only. No markdown. No chain-of-thought.\par

\end{flushleft}}

TEP: \path{{"diagnosis":"<one candidate>"}}. Bearings add confidence, per-candidate scores (−2…+2), and next\_action\_target. Lam uses the diagnosis-only object with its allowed labels.

Here, ‘No chain-of-thought’ is an output-format instruction: models were asked not to include a written rationale in the returned response. Provider-side reasoning effort was controlled separately, with reasoning traces excluded from the returned response.

\subsection*{C.2 External diagnosis sentences}

CWRU controlled advice

{\scriptsize\setlength{\baselineskip}{9.5pt}

\begin{flushleft}\ttfamily

An engineer who inspected the equipment suspects <true\_class>.\par

An engineer who inspected the equipment suspects <wrong\_class>.\par

\end{flushleft}}

Fitted-specialist prompts (CWRU, Paderborn, XJTU-SY, and TEP)

{\scriptsize\setlength{\baselineskip}{9.5pt}

\begin{flushleft}\ttfamily

CASE-SPECIFIC PRIOR (not itself physical evidence):\par

An external diagnostic advisor suggests: <label>.\par

\end{flushleft}}

TEP evidence-strength test

{\scriptsize\setlength{\baselineskip}{9.5pt}

\begin{flushleft}\ttfamily

CASE-SPECIFIC PRIOR (not itself physical evidence):\par

An engineer who inspected the process run suspects <wrong\_class>.\par

\end{flushleft}}

Lam Task A

{\scriptsize\setlength{\baselineskip}{9.5pt}

\begin{flushleft}\ttfamily

The process engineer believes this wafer run is normal and does not reflect a process/tool fault.\par

The process engineer believes this wafer run is abnormal and reflects a process/tool fault.\par

\end{flushleft}}

Lam Task B

{\scriptsize\setlength{\baselineskip}{9.5pt}

\begin{flushleft}\ttfamily

The process engineer suspects a TCP power fault.\par

The process engineer suspects a RF power fault.\par

The process engineer suspects a chamber-pressure fault.\par

The process engineer suspects a chlorine-flow fault.\par

The process engineer suspects a BCl3-flow fault.\par

\end{flushleft}}

Appendix Table C1 states the deterministic wrong-label rule used in each controlled-advice condition. Each case receives one incorrect label. The primary TEP integration condition is excluded because its external labels are predictions from the fitted specialist rather than deterministically assigned incorrect labels.

\Needspace{16\baselineskip}\emph{Appendix Table C1. Controlled wrong-advice label rules.}

\begin{table}[H]\centering\scriptsize\setlength{\tabcolsep}{2pt}

\begin{tabular}{>{\raggedright\arraybackslash}p{1.0in}>{\raggedright\arraybackslash}p{3.2in}>{\raggedright\arraybackslash}p{1.7in}}

\toprule

Condition & Deterministic rule & Depends on \\

\midrule

CWRU & Within each true class, the other three classes are assigned as evenly as possible after a seeded shuffle (seed 20260923). & True class and the stored map \\

Paderborn & Healthy $\rightarrow$ Outer-race damage; Inner-race damage $\rightarrow$ Healthy; Outer-race damage $\rightarrow$ Inner-race damage. & True class only \\

XJTU-SY & Among the other classes in the fixed candidate order, the label is chosen by case index. & True class, candidate order, and case index \\

TEP evidence-strength & NORMAL $\rightarrow$ IDV1 $\rightarrow$ IDV4 $\rightarrow$ IDV6 $\rightarrow$ IDV13 $\rightarrow$ IDV15 $\rightarrow$ NORMAL. & True class only \\

Lam 9600 Task B & Greedy balance over the other four classes, with ties broken by SHA-256 of wafer id, class, and seed 20260926. & True class, wafer id, and running counts \\

\bottomrule

\end{tabular}

\end{table}

\subsection*{C.3 Tennessee Eastman}

Prefix: “You are diagnosing one Tennessee Eastman Process run. Use only the evidence and the static reference. Do not invent numbers.”

The normal reference was deterministically constructed from the 50 training NORMAL runs using samples 21--500 and stored the mean and standard deviation for each process variable.

Static domain reference (identical in the primary integration and evidence-strength experiments; shown once):

{\scriptsize\setlength{\baselineskip}{9.5pt}

\begin{flushleft}\ttfamily

STATIC DOMAIN REFERENCE (identical every case; not evidence about this run):\par

VARIABLE ROLES (Downs–Vogel / Rieth names):\par

XMEAS(1..41): process measurements. XMV(1..11): manipulated variables.\par

SUBSYSTEM MEMBERSHIP:\par

- feed\_input\_streams: XMEAS(1,2,3,4,6), XMV(1,2,3,4)\par

[... 7 further lines ...]\par

PROCESS PATHWAY GROUPS (aggregates of subsystems; not a diagnosis):\par

- feed\_pathway: feed\_input\_streams\par

[... 3 further lines ...]\par

CANDIDATE FAULT DEFINITIONS (complete list; does not identify this case):\par

- NORMAL: Mode-1 normal operation, no disturbance.\par

- IDV1: A/C feed ratio / B composition constant (Stream 4).\par

- IDV4: Reactor cooling-water inlet temperature.\par

- IDV6: A-feed loss (Stream 1).\par

- IDV13: Slow drift in reaction kinetics.\par

- IDV15: Condenser cooling-water valve sticking.\par

Do not treat this reference as a ranking of the current run.\par

\end{flushleft}}

Same run, unaided then advised (reference omitted above; fixed per-case candidate order):

{\scriptsize\setlength{\baselineskip}{9.5pt}

\begin{flushleft}\ttfamily

\# Unaided\par

CASE EVIDENCE (deterministic z vs normal reference; window samples 21-500):\par

global: n\_|mean\_z|>=1=0 >=2=0 >=3=0 max|mean\_z|=0.186 median|mean\_z|=0.029 q75|mean\_z|=0.050\par

XMEAS(1) A Feed (Stream 1): mean\_z=+0.013 max\_|z|=2.728 slope=+0.0006 early\_late\_shift=+0.132\par

[... 51 further lines ...]\par

\vspace{2pt}

Choose exactly one candidate:\par

- IDV6\par

- IDV1\par

- IDV15\par

- IDV4\par

- IDV13\par

- NORMAL\par

\vspace{2pt}

Return JSON only: \{"diagnosis":"<candidate>"\}\par

No prose. No chain-of-thought. No confidence field.\par

\vspace{2pt}

\# Advised (identical case evidence; only the prior differs)\par

CASE EVIDENCE (deterministic z vs normal reference; window samples 21-500):\par

global: n\_|mean\_z|>=1=0 >=2=0 >=3=0 max|mean\_z|=0.186 median|mean\_z|=0.029 q75|mean\_z|=0.050\par

XMEAS(1) A Feed (Stream 1): mean\_z=+0.013 max\_|z|=2.728 slope=+0.0006 early\_late\_shift=+0.132\par

[... 51 further lines ...]\par

\vspace{2pt}

CASE-SPECIFIC PRIOR (not itself physical evidence):\par

An external diagnostic advisor suggests: IDV15.\par

\vspace{2pt}

Choose exactly one candidate:\par

- IDV6\par

- IDV1\par

- IDV15\par

- IDV4\par

- IDV13\par

- NORMAL\par

\vspace{2pt}

Return JSON only: \{"diagnosis":"<candidate>"\}\par

No prose. No chain-of-thought. No confidence field.\par

\end{flushleft}}

\subsection*{C.4 CWRU (opaque ID `C-000492CE`)}

Same case; shell/schema/candidate order shared; evidence blocks differ.

{\scriptsize\setlength{\baselineskip}{9.5pt}

\begin{flushleft}\ttfamily

\# Generic\par

You diagnose a rotating-machine bearing from summarized physical measurements.\par

Use only the evidence block. Do not invent numbers.\par

\vspace{2pt}

CASE ID: C-000492CE\par

Operating condition: 0 HP; 1796 RPM; drive-end @ 12000 Hz; window 4096 samples.\par

\vspace{2pt}

Observed evidence:\par

E1. Spectral band-energy shares: 0-500Hz=0.2\%, 500-1500Hz=0.6\%, 1500-3000Hz=25.3\%, 3000-6000Hz=73.9\%. Spectral entropy=0.647 (+11.1\% vs normal).\par

[... 5 further lines ...]\par

\vspace{2pt}

Choose the single most likely candidate.\par

Return VALID JSON ONLY:\par

\{"diagnosis":"<one candidate exactly>","confidence":<int 0-100>,"candidate\_scores":\{"<each candidate>":<int -2..2>\},"next\_action\_target":"<candidate or general>"\}\par

No prose. No chain-of-thought. Scores: +2 strong support ... -2 strong contradiction.\par

Candidates:\par

- Outer-race defect\par

- Inner-race defect\par

- Normal\par

- Ball defect\par

\vspace{2pt}

\# Task-aligned\par

You diagnose a rotating-machine bearing from summarized physical measurements.\par

Use only the evidence block. Do not invent numbers.\par

\vspace{2pt}

GENERAL DIAGNOSTIC REFERENCE (identical for every case; not a diagnosis):\par

- BPFO $\leftrightarrow$ outer-race; BPFI $\leftrightarrow$ inner-race; BSF $\leftrightarrow$ ball; FTF $\leftrightarrow$ cage.\par

[... 4 further lines ...]\par

\vspace{2pt}

CASE ID: C-000492CE\par

OPERATING CONDITION: 0 HP; 1796 RPM; shaft 29.93 Hz; drive-end @ 12000 Hz; window 4096;\par

theoretical BPFO=107.3, BPFI=162.1, BSF=70.5, FTF=11.9 Hz.\par

\vspace{2pt}

CURRENT PHYSICAL EVIDENCE:\par

E1. Envelope spectrum near BPFO (theoretical 107.3 Hz): peak SNR=23.52×; 2×BPFO harmonic SNR=19.31×.\par

[... 5 further lines ...]\par

\vspace{2pt}

Choose the single most likely candidate.\par

Return VALID JSON ONLY:\par

\{"diagnosis":"<one candidate exactly>","confidence":<int 0-100>,"candidate\_scores":\{"<each candidate>":<int -2..2>\},"next\_action\_target":"<candidate or general>"\}\par

No prose. No chain-of-thought. Scores: +2 strong support ... -2 strong contradiction.\par

Candidates:\par

- Outer-race defect\par

- Inner-race defect\par

- Normal\par

- Ball defect\par

\end{flushleft}}

\subsection*{C.5 Paderborn, XJTU, Lam}

On Paderborn, the generic view reports rotational speed, time-domain statistics, spectrum summaries, and band energy, while the aligned view adds geometry, frequency equations, and envelope SNR. The labels are Healthy, Inner-race damage, and Outer-race damage. XJTU follows the same design, with labels NORMAL, INNER\_RACE, and OUTER\_RACE; the rendered prompt bodies were not archived. Lam uses 19 standardized residuals. Task A labels are NORMAL and FAULTED\_PROCESS. Task B labels are TCP\_POWER, RF\_POWER, CHAMBER\_PRESSURE, CL2\_FLOW, and BCL3\_FLOW; the residual values are omitted here. The Lam 9600 source data are publicly downloadable from Eigenvector Research, but we do not redistribute them.

\subsection*{C.6 Decoding and reasoning settings}

\begin{table}[H]\centering\scriptsize\setlength{\tabcolsep}{3pt}

\begin{tabular}{p{2.05in}p{0.95in}p{0.72in}p{1.45in}p{0.7in}}

\toprule

Prompt family & Requested temperature & Seed & Reasoning effort / trace & Max tokens \\

\midrule

TEP integration, secondary sensitivity analyses, evidence-strength, and development & 0 & 20260928 & low / trace excluded & 512 \\

TEP sensitivity analysis with higher reasoning effort & 0 & 20260928 & high / trace excluded & 4096 \\

CWRU; Paderborn; XJTU & 0 & 20260924 & low or high / trace excluded & 2500 \\

Lam Tasks A and B & 0 & 20260926 & low / trace excluded & 64 \\

\bottomrule

\end{tabular}

\end{table}

Temperature 0 was requested; provider handling of sampling controls may differ. Four truncated Lam Sol calls were retried at max\_tokens 128; all other decoding settings were unchanged.

\section*{Appendix D. CWRU common-correct analysis}

Positive differences indicate reduced susceptibility to incorrect advice under task-aligned evidence. The restriction retains cases diagnosed correctly before advice under both presentations.

\Needspace{16\baselineskip}\emph{Appendix Table D1. CWRU common-correct susceptibility analysis.}

\begin{table}[H]\centering\scriptsize\setlength{\tabcolsep}{3pt}

\begin{tabular}{p{1.05in}p{0.7in}p{0.55in}p{0.7in}p{0.8in}p{1.65in}}

\toprule

Model & Common-correct n & Files & Generic & Task-aligned & Difference (95\% CI) \\

\midrule

GPT-5.6 Sol & 22 & 12 & 0.500 & 0.182 & 0.318 [0.067, 0.647] \\

Claude Sonnet 5 & 22 & 11 & 0.091 & 0.000 & 0.091 [0.000, 0.227] \\

Claude Opus 5.5 & 36 & 15 & 0.139 & 0.000 & 0.139 [0.045, 0.256] \\

Gemini 3.8 Flash & 23 & 12 & 0.348 & 0.087 & 0.261 [0.028, 0.647] \\

Qwen 3.8 & 27 & 12 & 0.667 & 0.185 & 0.481 [0.214, 0.731] \\

GLM-5.3 & 14 & 9 & 0.214 & 0.214 & 0.000 [−0.312, 0.308] \\

Kimi K3 & 14 & 9 & 0.429 & 0.000 & 0.429 [0.125, 0.733] \\

Seven-model mean, post hoc & N/A & 28 jointly resampled & 0.341 & 0.095 & 0.246 [0.105, 0.405] \\

\bottomrule

\end{tabular}

\end{table}

The final row is an unweighted summary across the fixed evaluated model panel. Its interval jointly resamples all 28 recording files across models; it supplements rather than replaces the model-specific results.

\section*{Appendix E. Evidence-strength analysis}

We tested whether a separately computed evidence-strength score predicted susceptibility to incorrect advice; the score was never shown to the LLM. On CWRU, the score was the leave-one-recording-file-out random-forest probability assigned to the true class under spectrum-only features. On TEP, the score was $p_{\mathrm{true}}-\max_{k\neq\mathrm{true}}p_k$ from the same fitted multinomial logistic-regression model used as the primary specialist. The TEP analysis was part of the prespecified held-out confirmation framework and used controlled incorrect advice with the deterministic wrong-label mapping in Table C1, rather than the specialist prediction. Odds ratios are reported per one-standard-deviation decrease in the dataset-specific score. The association was not statistically reliable on either dataset: CWRU OR 1.478, 95\% CI [0.789, 2.597] (206 eligible model–case rows); TEP OR 0.923, 95\% CI [0.746, 1.142] (688 model–case rows from 125 runs). Neither interval excluded 1. The TEP analysis conditions on initially correct unaided predictions; the exchangeable GEE did not return a finite fit, so the prespecified independence working correlation was used with the same mean model.

\section*{Appendix F. Reasoning-effort analyses}

\subsection*{F.1 CWRU reasoning-effort analysis}

Accuracy differences are high minus low. Susceptibility differences are low minus high conditional $H_{\mathrm{adj}}$ as defined in Section 3.8, so positive values mean lower susceptibility at the high reasoning-effort setting. Every interval includes zero.

\Needspace{16\baselineskip}\emph{Appendix Table F1. CWRU reasoning-effort sensitivity analysis.}

\begin{table}[H]\centering\scriptsize\setlength{\tabcolsep}{3pt}

\begin{tabular}{p{0.9in}p{0.68in}p{0.9in}p{1.05in}p{1.1in}p{1.15in}}

\toprule

Model & Parsed \(n\), low / high & Unaided accuracy, low to high & Accuracy difference (95\% CI) & Conditional $H_{\mathrm{adj}}$, low to high & Susceptibility difference (95\% CI) \\

\midrule

GPT-5.6 Sol & 72 / 72 & 0.347 to 0.333 & −0.014 [−0.062, 0.029] & 0.480 to 0.458 & 0.022 [−0.187, 0.319] \\

Claude Opus 5.5 & 72 / 72 & 0.583 to 0.639 & 0.056 [−0.055, 0.171] & 0.119 to 0.087 & 0.032 [−0.120, 0.199] \\

Gemini 3.8 Flash & 72 / 70 & 0.417 to 0.457 & 0.040 [−0.116, 0.199] & 0.333 to 0.188 & 0.146 [−0.091, 0.409] \\

Qwen 3.8 & 72 / 72 & 0.444 to 0.458 & 0.014 [−0.029, 0.060] & 0.719 to 0.818 & −0.099 [−0.253, 0.033] \\

GLM-5.3 & 72 / 72 & 0.375 to 0.417 & 0.042 [−0.101, 0.192] & 0.259 to 0.233 & 0.026 [−0.164, 0.222] \\

\bottomrule

\end{tabular}

\end{table}

Unaided accuracy uses successfully parsed responses. Conditional $H_{\mathrm{adj}}$ is computed on initially correct unaided rows under the Section 3.8 definition.

\subsection*{F.2 TEP sensitivity analysis with higher reasoning effort}

The primary held-out TEP evaluation called all seven models at the low reasoning-effort setting. This secondary analysis evaluates whether higher reasoning effort changes the integration deficit relative to the primary evaluation. It does not replace that seven-model result.

The selected four-model panel (Claude Opus 5.5, Gemini 3.8 Flash, GLM-5.3, and Kimi K3) was fixed from development diagnostics before any higher-reasoning confirmation outcome was examined. GPT-5.6 Sol, Claude Sonnet 5, and Qwen 3.8 were excluded because development diagnostics did not reliably distinguish their requested high setting from low. Decoding settings follow Appendix C.6; because the high setting also allowed a larger output-token limit, the manipulation is provider-specific and not separated from output length. Intervals are 10,000 percentile bootstrap resamples of the 150 confirmation runs, drawn jointly across the four models, with seed 20260924.

\Needspace{16\baselineskip}\emph{Appendix Table F2. Four-model TEP accuracies at low and high reasoning.}

\begin{table}[H]\centering\scriptsize\setlength{\tabcolsep}{3pt}

\begin{tabular}{p{1.15in}p{0.42in}p{0.62in}p{0.62in}p{0.72in}p{0.72in}p{0.62in}p{0.72in}}

\toprule

Model & n & Unaided low & Unaided high & Integrated low & Integrated high & Specialist & Median reasoning tokens (high) \\

\midrule

Claude Opus 5.5 & 150 & 0.833 & 0.827 & 0.833 & 0.827 & 0.833 & 0 \\

Gemini 3.8 Flash & 149 & 0.611 & 0.738 & 0.779 & 0.772 & 0.832 & 1,092 \\

GLM-5.3 & 150 & 0.547 & 0.627 & 0.720 & 0.773 & 0.833 & 90.5 \\

Kimi K3 & 150 & 0.627 & 0.633 & 0.747 & 0.793 & 0.833 & 478.5 \\

\bottomrule

\end{tabular}

\end{table}

Appendix Table F2 reports model-level accuracies. Effects were heterogeneous across models. Opus, which reported a median of zero reasoning tokens at the high setting, was essentially unchanged. Gemini’s unaided accuracy rose while its integrated accuracy did not. GLM-5.3 and Kimi K3 had higher integrated accuracy at the high setting. These point estimates are not separate confirmatory tests.

On the scored rows, unaided accuracy rose by 5.2 percentage points (95\% interval 2.2 to 8.2) and integrated accuracy by 2.2 points (−0.2 to 4.8). The difference of those changes was −3.0 points (−6.9 to 0.8). At the low setting, integrated output minus the better standalone source was −6.3 points (−13.5 to 1.0). At the high setting, the same contrast was −4.2 points (−10.5 to 0.8). The better source is recomputed inside each resample.

With the high-setting unaided label as the case-level reference (599 scored rows), 365 rows were correct for both sources, 134 for the specialist only, 58 for the LLM only, and 42 for neither. The integrated output missed 69 of 134 specialist corrections (51.5\%) and recorded 11 harmful acceptances in 58 LLM-only rows (19.0\%). The selector oracle was 93.0\%. On all 600 low-setting rows for the same four models, the corresponding counts were 335, 165, 57, and 43. Missed corrections were 84 of 165 (50.9\%) and harmful acceptances 6 of 57 (10.5\%), with a selector oracle of 92.8\%. The missed-correction asymmetry remained. The harmful-acceptance counts are small; no separate confidence interval was computed for them.

One Gemini unaided-high call was truncated and unparsed; analyses requiring the unaided-high label therefore use 599 rather than 600 rows. Coding the unparsed row as either correct or incorrect did not change the qualitative conclusion.

\section*{Appendix G. Bearing integration comparisons}

Appendix Table G1 records the six specialists used in the reported bearing integration comparisons. Accuracies and error counts are the stored out-of-fold values. The estimator is a 400-tree random forest in every row.

\Needspace{16\baselineskip}\emph{Appendix Table G1. Bearing specialist specifications.}

\begin{table}[H]\centering\scriptsize\setlength{\tabcolsep}{2pt}

\begin{tabular}{p{0.72in}p{0.72in}p{1.45in}p{1.15in}p{0.85in}p{0.72in}}

\toprule

Dataset & Feature set & Estimator & Out-of-fold unit & Out-of-fold accuracy & Specialist errors \\

\midrule

CWRU & strong & 400-tree random forest & recording file & 0.958 & 3 \\

CWRU & weak & 400-tree random forest & recording file & 0.917 & 6 \\

Paderborn & strong & 400-tree random forest & bearing & 0.724 & 24 \\

Paderborn & weak & 400-tree random forest & bearing & 0.437 & 49 \\

XJTU-SY & strong & 400-tree random forest & bearing & 0.909 & 8 \\

XJTU-SY & weak & 400-tree random forest & bearing & 0.761 & 21 \\

\bottomrule

\end{tabular}

\end{table}

Out-of-fold random forests use deterministic seed derivations from base seed 20260924.

The 22 included comparisons consist of the following dataset, evidence presentation, specialist, and model combinations (Appendix Table G2). Values are integrated accuracy minus the better standalone accuracy; the last field records whether the corresponding cluster interval is below, includes, or is above zero.

\Needspace{16\baselineskip}\emph{Appendix Table G2. Included bearing integration comparisons.}

\begin{table}[H]\centering\scriptsize\setlength{\tabcolsep}{3pt}

\begin{tabular}{p{1.65in}p{4.5in}}

\toprule

Dataset and condition & Model-level differences and interval classes \\

\midrule

Paderborn, task-aligned, strong specialist & Sol 0.000 (includes); Sonnet −0.093 (includes); Opus −0.069 (includes); Gemini −0.011 (includes); Qwen 0.069 (includes); GLM 0.046 (includes); Kimi −0.034 (includes) \\

Paderborn, task-aligned, weak specialist & Sol 0.023 (includes); Sonnet −0.084 (includes); Opus −0.023 (includes); Gemini 0.011 (includes); Qwen 0.034 (includes); GLM 0.000 (includes); Kimi −0.023 (includes) \\

XJTU, generic, weak specialist & Sonnet 0.000 (includes); Opus −0.250 (below); Qwen −0.011 (includes); Kimi −0.057 (includes) \\

XJTU, task-aligned, weak specialist & Sonnet −0.193 (below); Opus −0.136 (below); Gemini −0.023 (includes); Qwen 0.159 (above) \\

\bottomrule

\end{tabular}

\end{table}

Thus three intervals were below zero, eighteen included zero, and one was above zero. The CWRU specialists do not enter this 22-comparison inferential count because each made fewer than 15 errors under the predefined eligibility criteria: 3 on the strong feature set and 6 on the weak feature set. All 28 point estimates were below the better standalone source.

\section*{Appendix H. Paderborn and XJTU supporting analyses}

Paderborn and XJTU use related, but not identical, task-aligned interfaces. On Paderborn, model accuracy under the generic presentation was too low for a clean comparison of susceptibility to incorrect advice. The task-aligned presentation raised diagnostic accuracy and reduced the all-case harmful-switching rate, so the two changes cannot be separated cleanly. A supporting Sol common-correct contrast was 0.758 [0.538, 0.917] on 33 cases from 13 bearings.

On XJTU, task-aligned evidence lowered the all-case harmful-switching rate for all seven evaluated models, while balanced diagnostic accuracy also increased. For several models, few cases were correct under both presentations because only 11 physical bearings were available. These results are consistent with the bearing-domain evidence-presentation finding, while diagnostic competence and advice-related switching change together and CWRU remains the cleaner fixed-case analysis. This appendix reports balanced accuracy for the seven-model supporting analysis, whereas Table N1 reports ordinary multiclass accuracy for the post-hoc panel. A post-hoc decomposition of normal-versus-fault errors is reported in Appendix N.

\section*{Appendix I. TEP development presentation contrast}

A separate TEP presentation contrast reorganized the same semantic process measurements by subsystem and pathway without adding a new physical transform. Retest-adjusted harmful switching was 0.107 under the generic presentation and 0.124 under the subsystem presentation, a difference of −0.017 (95\% CI [−0.036, 0.003]). This development-set analysis is separate from the held-out TEP confirmation evaluation.

\section*{Appendix J. Real-equipment advice-vulnerability check}

In the Lam five-class task, 23 of 60 unaided model–case judgments (three models by 20 wafers) were correct. Six were changed to the incorrect external label, across four wafers, while their paired repeats did not flip. The result is descriptive.

\section*{Appendix K. TEP secondary sensitivity analyses}

This appendix reports the three TEP secondary sensitivity analyses defined in Section 3.5.

The arbitration condition retained the logistic specialist and added exactly:

\noindent\begin{minipage}{\linewidth}
{\scriptsize\setlength{\baselineskip}{9.5pt}

\begin{flushleft}\ttfamily

You have two fallible sources: your own reading of the case\par

evidence above and the external diagnostic model's prediction.\par

Decide which diagnosis is better supported by the case evidence.\par

Do not automatically defer to either source.\par

\end{flushleft}}
\end{minipage}

\Needspace{20\baselineskip}Appendix Table K1 reports the paired model-level implicit and explicit-arbitration results.

\Needspace{18\baselineskip}\emph{Appendix Table K1. Primary implicit and arbitration-condition integrated accuracy by model.}

\begin{table}[H]\centering\scriptsize\setlength{\tabcolsep}{3pt}

\begin{tabular}{>{\raggedright\arraybackslash}p{1.35in}>{\raggedright\arraybackslash}p{0.95in}>{\raggedright\arraybackslash}p{0.95in}>{\raggedright\arraybackslash}p{1.55in}}

\toprule

Model & Implicit integration & Arbitration condition & Arbitration − implicit \\

\midrule

GPT-5.6 Sol & 0.820 & 0.813 & −0.007 \\

Claude Sonnet 5 & 0.840 & 0.793 & −0.047 \\

Claude Opus 5.5 & 0.833 & 0.833 & 0.000 \\

Gemini 3.8 Flash & 0.780 & 0.773 & −0.007 \\

Qwen 3.8 & 0.680 & 0.673 & −0.007 \\

GLM-5.3 & 0.720 & 0.707 & −0.013 \\

Kimi K3 & 0.747 & 0.733 & −0.013 \\

Seven-model mean & 0.774 & 0.761 & −0.013 [−0.031, +0.005] \\

\bottomrule

\end{tabular}

\end{table}

Under the same error definition as Figure 3, the arbitration condition had 161/295 missed corrections and 6/99 harmful acceptances; five of the six harmful acceptances copied the specialist’s wrong label.

The alternative-specialist condition used the alternative 400-tree random forest from the fixed development candidate set. Both candidates passed the same eligibility criteria; logistic regression was selected for the primary test by the prespecified ranking rule because it provided more pooled LLM-only-correct rows. Confirmation outcomes played no role in either choice.

\Needspace{20\baselineskip}Appendix Table K2 reports the model-level alternative-specialist results.

\Needspace{18\baselineskip}\emph{Appendix Table K2. Alternative-specialist implicit-integration results by model.}

\begin{table}[H]\centering\scriptsize\setlength{\tabcolsep}{3pt}

\begin{tabular}{p{0.85in}p{0.95in}p{0.95in}p{1.35in}p{1.45in}}

\toprule

Model & Parsed n & Integrated & RF specialist & Selector oracle \\

\midrule

GPT-5.6 Sol & 150 & 0.827 & 0.840 & 0.913 \\

Claude Sonnet 5 & 150 & 0.833 & 0.840 & 0.900 \\

Claude Opus 5.5 & 150 & 0.833 & 0.840 & 0.913 \\

Gemini 3.8 Flash & 150 & 0.773 & 0.840 & 0.900 \\

Qwen 3.8 & 150 & 0.673 & 0.840 & 0.913 \\

GLM-5.3 & 147 & 0.755 & 0.837 & 0.905 \\

Kimi K3 & 150 & 0.780 & 0.840 & 0.913 \\

Seven-model mean & 1,047 total & 0.782 & 0.840 & 0.908 \\

\bottomrule

\end{tabular}

\end{table}

Three GLM-5.3 responses in the alternative-specialist condition failed to parse. Extreme coding of those responses as all correct or all incorrect left the integration deficit below zero; the qualitative conclusion was unchanged.

The prior-label condition, a post hoc sensitivity analysis, kept the arbitration text and inserted one line; the specialist prediction precedes that block.

\noindent\begin{minipage}{\linewidth}
{\scriptsize\setlength{\baselineskip}{9.5pt}

\begin{flushleft}\ttfamily

An external diagnostic advisor suggests: <specialist>.\par

\par

You have two fallible sources: your own reading of the case\par

evidence above and the external diagnostic model's prediction.\par

\par

Your earlier diagnosis from the case evidence alone: <z0>.\par

\par

Decide which diagnosis is better supported by the case evidence.\par

Do not automatically defer to either source.\par

\end{flushleft}}
\end{minipage}

Appendix Table K3 reports the model-level prior-label accuracies.

\Needspace{18\baselineskip}\emph{Appendix Table K3. Prior-label integrated accuracy by model.}

\begin{table}[H]\centering\scriptsize\setlength{\tabcolsep}{2pt}

\begin{tabular}{>{\raggedright\arraybackslash}p{1.35in}>{\raggedright\arraybackslash}p{0.95in}>{\raggedright\arraybackslash}p{0.95in}>{\raggedright\arraybackslash}p{1.55in}}

\toprule

Model & Arbitration & Prior-label & Prior-label − arbitration \\

\midrule

GPT-5.6 Sol & 0.813 & 0.827 & +0.013 \\

Claude Sonnet 5 & 0.793 & 0.653 & −0.140 \\

Claude Opus 5.5 & 0.833 & 0.833 & 0.000 \\

Gemini 3.8 Flash & 0.773 & 0.673 & −0.100 \\

Qwen 3.8 & 0.673 & 0.667 & −0.007 \\

GLM-5.3 & 0.707 & 0.673 & −0.033 \\

Kimi K3 & 0.733 & 0.727 & −0.007 \\

Seven-model mean & 0.761 & 0.722 & −0.039 [−0.058, −0.020] \\

\bottomrule

\end{tabular}

\end{table}

\section*{Appendix L. Advice vulnerability across datasets}

Figure L1 summarizes advice vulnerability across the five datasets. Each point conditions on an initially correct unaided judgment and an incorrect external diagnosis. Paired-repeat retention measures ordinary instability; advised retention measures the corresponding outcome after conflicting external information.

\begin{figure}[H]

\centering

\includegraphics[width=0.95\textwidth]{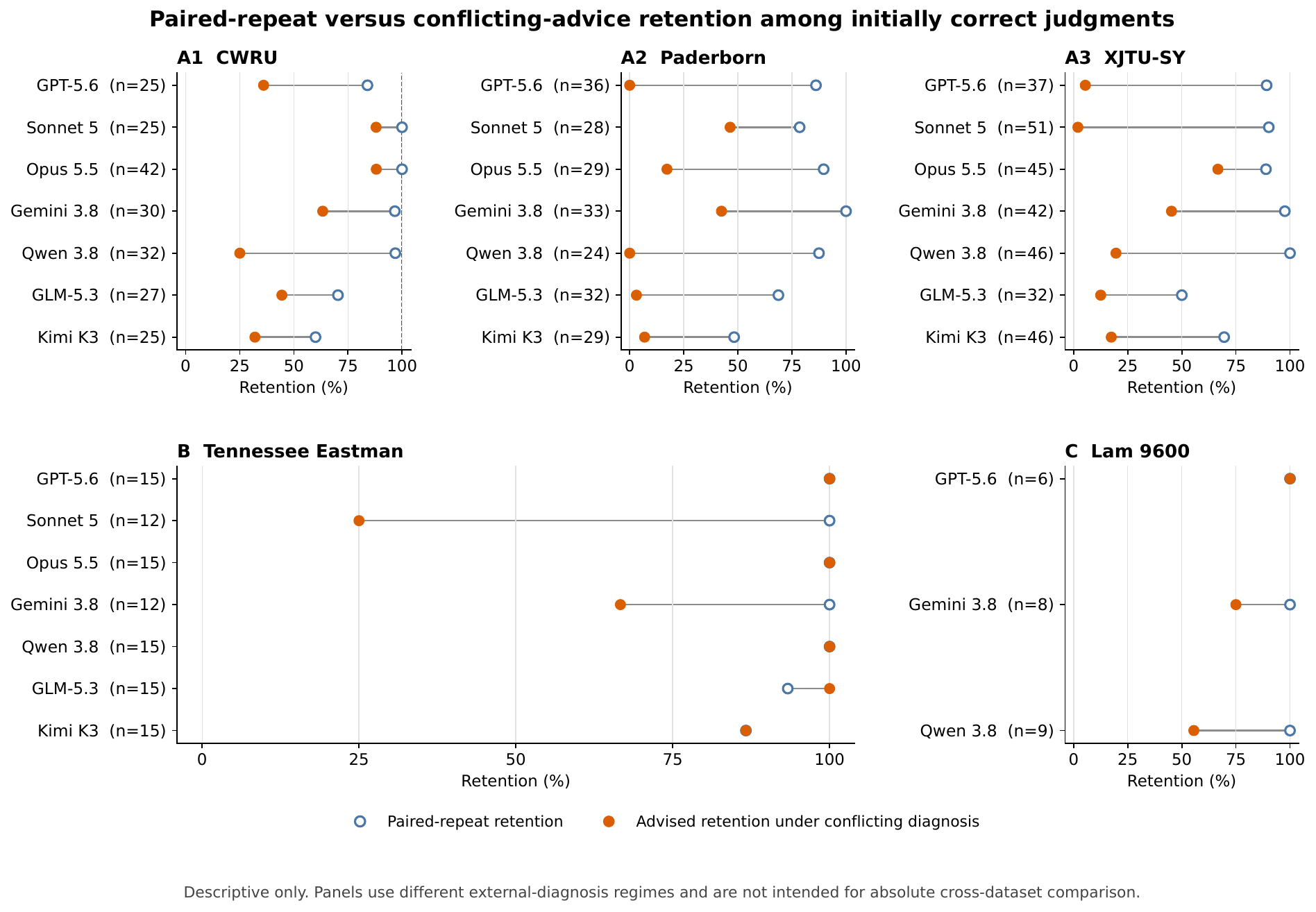}

\caption*{\textbf{Appendix Figure L1.} Advice vulnerability across five diagnostic datasets. Panel A shows fixed generic controlled wrong-advice conditions for CWRU, Paderborn, and XJTU-SY. Panel B shows conflicts with the primary logistic-regression specialist on the Tennessee Eastman confirmation set; the arbitration, prior-label, and alternative-specialist conditions are excluded. Panel C shows the five-class Lam 9600 controlled-advice check. Open markers show paired-repeat retention, filled markers show advised retention, and each model label gives the eligible denominator. Within each panel, the difference between paired-repeat retention and advised retention shows the loss of initially correct judgments associated with conflicting external diagnostic information beyond ordinary repeat instability. Panels use different external-diagnosis regimes; no cross-dataset pooling or inference is performed.}

\end{figure}

\clearpage
\section*{Appendix M. Post-hoc TEP NORMAL-class source comparison}

To clarify the source-reliance trade-off discussed in Section 5.2, we performed a post-hoc analysis restricted to the 25 true-NORMAL runs in the primary Tennessee Eastman confirmation set, using only stored predictions. We define the normal-to-fault error rate (binary false-alarm rate) as the share of cases with true label NORMAL whose predicted label is any of IDV1, IDV4, IDV6, IDV13, or IDV15. This conditional classification error should not be interpreted as an operational plant alarm frequency.

\Needspace{14\baselineskip}\emph{Appendix Table M1. NORMAL-to-fault rates on the 25 true-NORMAL TEP confirmation runs.}

\begin{table}[H]\centering\scriptsize\setlength{\tabcolsep}{4pt}

\begin{tabular}{p{1.5in}p{1.15in}p{1.0in}p{1.55in}}

\toprule

Source & Unit & NORMAL$\rightarrow$fault & Rate \\

\midrule

Unaided LLM & 175 model–runs & 8/175 & 4.6\% [0.6, 9.7] \\

Specialist & 25 runs & 14/25 & 56.0\% [36.0, 76.0] \\

Integrated output & 175 model–runs & 17/175 & 9.7\% [5.7, 14.3] \\

\bottomrule

\end{tabular}

\end{table}

Intervals resample the 25 NORMAL runs jointly across models; the specialist prediction is counted once per run.

Relative to the unaided LLM, the integrated output differed by $+5.1$ percentage points (95\% interval $-1.7$ to $11.4$). Relative to the selected TEP specialist, it differed by $-46.3$ points ($-63.4$ to $-29.7$). The integrated-versus-unaided interval includes zero, so these data do not establish that integration increased NORMAL-to-fault errors. The integrated output had a substantially lower NORMAL-to-fault error rate than this specialist.

When the unaided LLM correctly predicted NORMAL but the specialist predicted a fault, integration preserved NORMAL in 80 of 94 model–run cases (85.1\%) and predicted the specialist’s fault label in 14 (14.9\%). All 14 adopted false alarms exactly copied the specialist’s IDV15 prediction. Overall, 14 of the 17 integrated NORMAL-to-fault errors correspond to the harmful-acceptance cases already counted in Figure 3; the remaining three arose under other source-correctness configurations. Harmful acceptance conditions on an initially correct unaided prediction paired with an incorrect specialist label, whereas the NORMAL-to-fault rate conditions on true NORMAL, so the denominators differ.

\section*{Appendix N. Post-hoc normal-versus-fault error profiles}

To examine whether changes in diagnostic competence were accompanied by changes in error composition, we performed a post-hoc analysis of stored unaided predictions on the three bearing datasets with explicit normal or healthy classes. We report normal/healthy-to-fault errors (binary false-alarm rate) separately from fault-to-normal/healthy misses. These conditional classification errors should not be interpreted as operational plant alarm frequencies. The generic and task-aligned presentations do not contain identical information; the task-aligned condition adds fault-relevant physical structure. The analysis therefore characterizes the evaluated interfaces rather than isolating a formatting effect. Dataset panels follow the stored post-hoc definitions and are not pooled. The Paderborn panel combines outputs from two evaluation rounds and was not fixed as a single seven-model panel.

\Needspace{16\baselineskip}\emph{Appendix Table N1. Post-hoc normal/healthy-versus-fault error profiles under generic and task-aligned evidence.}

\begin{table}[H]\centering\scriptsize\setlength{\tabcolsep}{2.5pt}

\begin{tabular}{p{0.72in}p{0.72in}p{0.72in}p{1.15in}p{0.72in}p{0.72in}p{1.05in}}

\toprule

Dataset & Generic N$\rightarrow$F & Aligned N$\rightarrow$F & Paired N$\rightarrow$F diff. & Generic F$\rightarrow$N & Aligned F$\rightarrow$N & Overall multiclass acc.\ (gen.\ $\rightarrow$ al.) \\

\midrule

CWRU & 22.2\% & 34.9\% & $+12.7$ pp [$-15.9$, $41.3$] & 11.1\% & 6.9\% & 40.9\% $\rightarrow$ 66.1\% \\

Paderborn & 88.1\% & 94.8\% & $+7.1$ pp [$-0.8$, $14.3$] & 18.8\% & 0.2\% & 34.6\% $\rightarrow$ 70.3\% \\

XJTU-SY & 26.1\% & 87.5\% & $+61.4$ pp [$50.0$, $73.4$] & 48.3\% & 0.6\% & 50.0\% $\rightarrow$ 51.4\% \\

\bottomrule

\end{tabular}

\end{table}

N$\rightarrow$F denotes normal/healthy-to-fault error; F$\rightarrow$N denotes fault-to-normal/healthy miss. Overall multiclass accuracy is ordinary accuracy on the post-hoc panel and is not balanced accuracy. Intervals are paired cluster bootstraps for the ALIGNED$-$GENERIC difference in normal/healthy-to-fault error.

The point estimates do not support a general claim that task-aligned evidence reduces normal-to-fault errors. CWRU and Paderborn showed positive but statistically inconclusive paired differences, whereas XJTU-SY showed a large increase accompanied by an almost complete disappearance of fault-to-NORMAL misses. These post-hoc results are better interpreted as dataset-dependent shifts in the diagnostic operating profile than as a uniform benefit or cost of task-aligned evidence. This complements the main intervention result: reduced susceptibility to incorrect advice does not imply that every component of diagnostic performance improves simultaneously.

\section*{Appendix O. Computational and API budget}

\begin{table}[H]\centering\footnotesize\setlength{\tabcolsep}{4pt}

\begin{tabular}{p{2.6in}p{1.6in}}

\toprule

Study group & Approx.\ cost (USD) \\

\midrule

Bearing studies & 271 \\

Tennessee Eastman studies & 152 \\

Lam 9600 analysis & 3 \\

Total & 426 \\

\bottomrule

\end{tabular}

\end{table}

LLM inference was performed through the OpenRouter API. Approximate API expenditure was \$426 in total, including retries associated with transient API or provider errors.

\section*{Appendix P. Data and code availability}

Analysis code and reproduction artifacts are not yet publicly available. Public datasets are available from their original maintainers: the CWRU \hyperref[cite:cwru_bearing]{Bearing Data Center}, \hyperref[cite:lessmeier2016]{Lessmeier et al. (2016)}, \hyperref[cite:wang2020]{Wang et al. (2020)}, \hyperref[cite:rieth2017]{Rieth et al. (2017)}, and the Eigenvector Research distribution of the Lam 9600 data described by \hyperref[cite:wise1999]{Wise et al. (1999)}. The Lam distribution page is \path{https://eigenvector.com/resources/data-sets/metal-etch-data-for-fault-detection-evaluation/}. We do not redistribute the Lam source files.

\end{document}